\documentclass{article}
\PassOptionsToPackage{numbers, compress}{natbib}

\newif\ifpreprint\preprintfalse

\ifpreprint
    \usepackage[preprint]{tackling_climate_workshop_style}
    \newcommand{\ghlink}{https://github.com/joeloskarsson/neural-lam}
\else
    \usepackage[final]{tackling_climate_workshop_style}
    \newcommand{\ghlink}{https://github.com/joeloskarsson/neural-lam}
\fi

\usepackage[utf8]{inputenc} %
\usepackage[T1]{fontenc}    %
\usepackage{hyperref}       %
\usepackage{url}            %
\usepackage{booktabs}       %
\usepackage{amsfonts}       %
\usepackage{nicefrac}       %
\usepackage{microtype}      %

\title{\papertitle}

\usepackage[utf8]{inputenc}
\usepackage{paralist} %
\usepackage{xspace}
\usepackage{graphicx}
\usepackage{mathtools}
\usepackage{cleveref}
\creflabelformat{equation}{#2#1#3} %
\crefrangelabelformat{equation}{#3#1#4--#5\crefstripprefix{#1}{#2}#6}
\crefrangeformat{subfigure}{(#3\onlyletter{#1}#4--#5\onlyletter{#2}#6)}
\usepackage{siunitx}
\usepackage{multibib}
\newcites{app}{Appendix References}

\usepackage{xcolor}
\definecolor{LiUblue}{RGB}{0,185,231}
\definecolor{LiUorange}{RGB}{255,100,66}
\definecolor{LiUgreen}{RGB}{0,207,181}
\usepackage{caption}
\usepackage{subcaption}
\usepackage{tikz}

\let\originalleft\left
\let\originalright\right
\renewcommand{\left}{\mathopen{}\mathclose\bgroup\originalleft}
\renewcommand{\right}{\aftergroup\egroup\originalright}

\usepackage[acronyms, shortcuts]{glossaries}
\glsdisablehyper
\newacronym{RNN}{RNN}{Recurrent Neural Network}
\newacronym{CNN}{CNN}{Convolutional Neural Network}
\newacronym{GNN}{GNN}{Graph Neural Network}
\newacronym{MLP}{MLP}{Multi-Layer Perceptron}
\newacronym{MPNN}{MPNN}{Message Passing Neural Network}
\newacronym{GRU}{GRU}{Gated Recurrent Unit}
\newacronym{ELBO}{ELBO}{Evidence Lower Bound}

\newacronym{RMSE}{RMSE}{Root Mean Squared Error}
\newacronym{MSE}{MSE}{Mean Squared Error}
\newacronym{MAE}{MAE}{Mean Absolute Error}

\newacronym{ODE}{ODE}{Ordinary Differential Equation}
\newacronym{PDE}{PDE}{Partial Differential Equation}
\newacronym{GMRF}{GMRF}{Gaussian Markov Random Field}

\newacronym{NWP}{NWP}{Numerical Weather Prediction}
\newacronym{NeurWP}{NeurWP}{Neural Weather Prediction}
\newacronym{LAM}{LAM}{Limited Area Model}
\newacronym{MetCoOp}{MetCoOp}{Meterological Cooperation on Operational NWP}
\newacronym{SMHI}{SMHI}{Swedish Meteorological and Hydrological Institute}
\newacronym{MEPS}{MEPS}{MetCoOp Ensemble Prediction System}

\usepackage{amssymb} %
\usepackage{amsmath}
\usepackage{bm} %

\newcommand{\eq}[2][my_equation]{\begin{equation}\label{eq:#1}#2\end{equation}}
\newcommand{\al}[2][my_equation]{\begin{align}\label{eq:#1}#2\end{align}}

\newcommand{\loss}{\mathcal{L}}

\newcommand{\set}[1]{\left\{ #1 \right\}}
\newcommand{\setsize}[1]{\left| #1 \right|}

\renewcommand{\b}[1]{\bm{#1}} %

\newcommand{\indicator}[1]{\mathbb{I}_{\left\{ #1 \right\}}}

\newcommand{\papertitle}{Graph-based Neural Weather Prediction for Limited Area Modeling}

\newcommand{\citeauth}[1]{\mbox{\citeauthor{#1}~\cite{#1}}}

\newcommand{\alse}[2][my_eq]{\begin{subequations}\label{eq:#1}\begin{align}#2\end{align}\end{subequations}}

\newcommand{\wvar}[1]{\texttt{#1}}%

\newcommand{\wstate}{X}

\newcommand{\pred}{\hat{\wstate}}

\newcommand{\timei}{t}

\newcommand{\arfunc}{f}
\newcommand{\modelfunc}{\hat{\arfunc}}
\newcommand{\arorder}{p}
\newcommand{\playerindex}{k}

\newcommand{\ngridpoints}{N}
\newcommand{\statedim}{S}
\newcommand{\samplelength}{T}
\newcommand{\meshdist}{d_m}

\newcommand{\gc}{GraphCast\xspace}
\newcommand{\ms}{GC-LAM\xspace}
\newcommand{\ol}{1L-LAM\xspace}
\newcommand{\hi}{Hi-LAM\xspace}

\newcommand{\gtm}{\text{G2M}}
\newcommand{\mtm}{\text{M2M}}
\newcommand{\mtg}{\text{M2G}}

\newcommand{\sameedge}[1]{[#1\,\leftrightarrow\,#1]}
\newcommand{\upedge}[2]{[#1 \,\nearrow\, #2]}
\newcommand{\downedge}[2]{[#1 \,\searrow\, #2]}

\newcommand{\himaxl}{L}

\newcommand{\boundary}{\mathbb{B}}
\newcommand{\interior}{\mathbb{G}}
\newcommand{\edgeset}{\mathcal{E}}
\newcommand{\edges}[1]{\edgeset{}^{#1}}
\newcommand{\mtmedges}{\edges{\mtm}}
\newcommand{\gtmedges}{\edges{\gtm}}
\newcommand{\mtgedges}{\edges{\mtg}}

\newcommand{\msameedges}[1]{\edges{\sameedge{#1}}}
\newcommand{\mupedges}[2]{\edges{\upedge{#1}{#2}}}
\newcommand{\mdownedges}[2]{\edges{\downedge{#1}{#2}}}

\newcommand{\nodevec}[1]{\b{v}_{#1}}
\newcommand{\gridnodevec}[1]{\b{v}_{#1}^{G}}
\newcommand{\meshnodevec}[1]{\b{v}_{#1}^{M}}
\newcommand{\levelnodevec}[2]{\b{v}_{#2}^{M[#1]}}

\newcommand{\edgevec}[3]{\b{e}_{#2 \rightarrow #3}^{#1}}
\newcommand{\mtmvec}[2]{\edgevec{\mtm}{#1}{#2}}
\newcommand{\gtmvec}[2]{\edgevec{\gtm}{#1}{#2}}
\newcommand{\mtgvec}[2]{\edgevec{\mtg}{#1}{#2}}

\newcommand{\edgevecprime}[3]{\b{\tilde{e}}_{#2 \rightarrow #3}^{#1}}
\newcommand{\mtmvecprime}[2]{\edgevecprime{\mtm}{#1}{#2}}
\newcommand{\gtmvecprime}[2]{\edgevecprime{\gtm}{#1}{#2}}
\newcommand{\mtgvecprime}[2]{\edgevecprime{\mtg}{#1}{#2}}

\newcommand{\msamevec}[3]{\edgevec{\sameedge{#1}}{#2}{#3}}
\newcommand{\mupvec}[4]{\edgevec{\upedge{#1}{#2}}{#3}{#4}}
\newcommand{\mdownvec}[4]{\edgevec{\downedge{#1}{#2}}{#3}{#4}}
\newcommand{\msamevecprime}[3]{\edgevecprime{\sameedge{#1}}{#2}{#3}}
\newcommand{\mupvecprime}[4]{\edgevecprime{\upedge{#1}{#2}}{#3}{#4}}
\newcommand{\mdownvecprime}[4]{\edgevecprime{\downedge{#1}{#2}}{#3}{#4}}

\newcommand{\mlpfunc}[2]{\text{MLP}^{#1}_{#2}}
\newcommand{\mlp}[3]{\mlpfunc{#1}{#2}\left(#3\right)}
\newcommand{\gtmmlp}[2]{\mlp{\gtm}{#1}{#2}}
\newcommand{\mtmmlp}[2]{\mlp{\mtm}{#1}{#2}}
\newcommand{\mtgmlp}[2]{\mlp{\mtg}{#1}{#2}}

\newcommand{\msamemlp}[3]{\mlp{\sameedge{#1}}{#2}{#3}}
\newcommand{\mupmlp}[4]{\mlp{\upedge{#1}{#2}}{#3}{#4}}
\newcommand{\mdownmlp}[4]{\mlp{\downedge{#1}{#2}}{#3}{#4}}

\newcommand{\mupdatemlp}[3]{\mlp{M[#1]}{#2}{#3}}

\newcommand{\readout}{\text{Dec.}}
\newcommand{\readoutnodemlp}[2]{\mupdatemlp{#1}{\readout}{#2}}
\newcommand{\readoutedgemlp}[3]{\mdownmlp{#1}{#2}{\readout}{#3}}

\newcommand{\init}{\text{Enc.}}
\newcommand{\initnodemlp}[2]{\mupdatemlp{#1}{\init}{#2}}
\newcommand{\initedgemlp}[3]{\mupmlp{#1}{#2}{\init}{#3}}

\newcommand{\timeprogress}{\tau}
\newcommand{\timelength}{\mathcal{T}}

\newcommand{\rolloutlen}{N_\text{Rollout}}
\newcommand{\inittime}{t_{\text{Init}}}
\newcommand{\stateweight}{\omega}
\newcommand{\invdiffweight}{\lambda}

\author{%
Joel Oskarsson\\
Link\"{o}ping University\\
Link\"{o}ping, Sweden\\
\texttt{joel.oskarsson@liu.se}
\And
Tomas Landelius\\
SMHI\\
Norrk\"{o}ping, Sweden\\
\texttt{tomas.landelius@smhi.se}
\And
Fredrik Lindsten\\
Link\"{o}ping University\\
Link\"{o}ping, Sweden\\
\texttt{fredrik.lindsten@liu.se}
}

\begin{document}

\maketitle

\begin{abstract}
The rise of accurate machine learning methods for weather forecasting is creating radical new possibilities for modeling the atmosphere.
In the time of climate change, having access to high-resolution forecasts from models like these is also becoming increasingly vital.
While most existing \acrfull{NeurWP} methods focus on global forecasting, an important question is how these techniques can be applied to limited area modeling.
In this work we adapt the graph-based \acrshort{NeurWP} approach to the limited area setting and propose a multi-scale hierarchical model extension.
Our approach is validated by experiments with a local model for the Nordic region.
\end{abstract}

\section{Introduction}
Accurately forecasting weather is an immense challenge, but also a problem with broad impact on society.
Today, \gls{NWP} systems combine vast amounts of physics knowledge with powerful computational resources in order to model the atmosphere. %
Performing these computations is however time-consuming, limiting the possibility to model finer resolutions and detect rare events \cite{extreme_weather_forecasting}.
Forecasting these events is vital in adapting society to the effects of climate change \cite{ipcc_2023}.
Recently, data-driven machine learning models have shown impressive weather forecasting performance, matching or even outperforming existing \gls{NWP} systems \cite{rise_of_neurwp, panguweather, graphcast}.
These \gls{NeurWP} models produce forecasts in a fraction of the time of traditional \gls{NWP}, opening up many new possibilities.
One successful family of \gls{NeurWP} models, that we focus on in this work, utilize \glspl{GNN} to produce forecasts \cite{keisler, graphcast}.
As these models are trained on \gls{NWP} data, with built-in physics assumptions, they represent a useful blend of existing knowledge and new possibilities introduced by machine learning.

While most existing work in \gls{NeurWP} has been focused on global weather forecasting, it is of substantial interest how these methods can be utilized also for \glspl{LAM}.
Such local models are used by many institutes to create high resolution forecasts for specific regions of interest \cite{arome_metcoop,lam_usa,lam_uk,lam_france,lam_germany,lam_india,lam_china,lam_japan}.
In this paper we adapt graph-based \gls{NeurWP} to the limited area setting, and also introduce a hierarchical model extension.\footnote{Our code is available at \url{\ghlink}.}
The graph-based framework is an attractive choice for \glspl{LAM} due to the freedom in designing the associated graphs.
By choosing these graphs appropriately, a general model formulation can be applied to regions of different shape or alignment with global coordinate systems.
We evaluate our graph-based models on a dataset from the \gls{MEPS} \gls{LAM} \cite{arome_metcoop}.

\paragraph{Climate Impact}
Due to climate change, access to efficient and accurate weather modeling is increasingly important.
Fast \gls{NeurWP} \glspl{LAM} contribute to climate change mitigation and adaptation through:
\begin{inparaenum}[1)]
    \item Extreme weather prediction, made possible by fast \gls{NeurWP} models enabling large ensemble forecasts.
    Detecting extreme events requires modeling the full distribution of weather states, which is typically done by creating an ensemble of forecasts from perturbed versions of initial states \cite{fundamentals_of_nwp}.
    \item Providing accurate forecasts for renewable energy production, enabling better integration of these volatile sources into existing energy systems.
    \item Reducing the energy required to produce fine-grained weather forecasts \cite{fourcastnet}.
\end{inparaenum}
In \cref{sec:climate_impact_app} we discuss paths to climate impact in more detail.

\section{Related Work}
\paragraph{Global \gls{NeurWP}}
Our work builds on graph-based \gls{NeurWP} \cite{keisler} and in particular the \gc model \cite{graphcast}.
Apart from this approach, another family of global \gls{NeurWP} models rely on transformer architectures \cite{aiayn, swin_transformer}, treating spatial or spatio-temporal patches of the weather state as input tokens \cite{panguweather, fengwu, fourcastnet, climax, atmorep}.
Some of these transformer-based approaches have been framed as \textit{foundation models}, aiming to learn general representations of the atmosphere that can be utilized in diverse downstream tasks \cite{climax, atmorep}.
Such models can be fine-tuned for limited area forecasting, but using the same variables as the global models and an area aligned with the global latitude-longitude grid.
Other noteworthy models are based on \textit{neural operator} architectures \cite{fouriernop,afno}, which perform efficient global convolutions by operating in the frequency domain \cite{fourcastnet, sphericalfno}.

\paragraph{Hierarchical Graph Neural Networks}
Our proposed hierarchical \gls{GNN} structure shares many ideas with the models of \citeauth{multiscalemgn} and \citeauth{simulating_cont_dynamics}.
While also motivated by capturing multiple spatial scales, they apply hierarchical \glspl{GNN} to general \gls{PDE} solving, rather than the weather forecasting task.

\section{Graph-based Forecasting for a Limited Area}
\newcommand{\tikzcircle}[2][LiUblue,fill=LiUblue]{\tikz\draw[#1,radius=#2] (0,0) circle ;}%
\begin{figure}[t] %
    \centering
    \includegraphics[width=\linewidth]{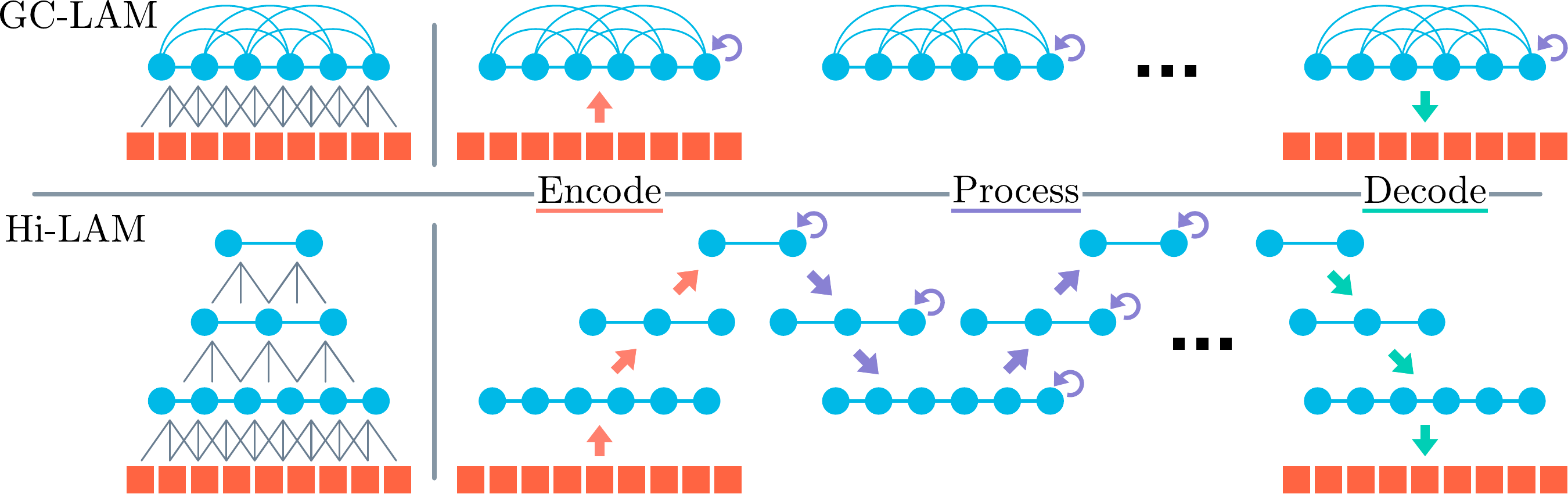}
    \caption{Overview of the prediction process of the \ms and \hi models. Inputs at grid nodes \textcolor{LiUorange}{$\blacksquare$} are encoded to mesh nodes \tikzcircle{3pt}, processed and decoded back to produce a one step prediction.}
    \label{fig:model_overview}
\end{figure}

The vast majority of \gls{NeurWP} models learn to approximate a function
$\wstate^{\timei} = \arfunc\left(\wstate^{\timei - 1}, \dots, \wstate^{\timei - p}\right)$
that autoregressively maps from the $\arorder$ last weather states to the state $\wstate^{\timei}$ at time step $\timei$ \cite{panguweather, keisler, fourcastnet}.
Each such weather state $\wstate^\timei$ is a matrix of shape $\ngridpoints \times \statedim$ containing $S$ different atmospheric variables at $N$ different spatial locations.
Each atmospheric variable is typically modeled at a number of vertical levels, but for clarity we here consider all levels for all variables stacked as one vector of length $\statedim$.
We refer to the $N$ spatial locations as \textit{grid cells} or (in the graph context) as \textit{grid nodes}.
In global models each such grid node typically corresponds to a cell in a latitude-longitude grid~\cite{panguweather, graphcast}.
Using machine learning, a mapping $\modelfunc \approx \arfunc$ can be learned from a dataset of weather state trajectories $\wstate^{1}, \wstate^{2}, \dots, \wstate^{\samplelength}$.
Given $\arorder$ initial states, a full forecast can be produced by unrolling the model multiple time steps, applying $\modelfunc$ iteratively to its own predictions.

\paragraph{Graph-based Forecasting}
In graph-based \gls{NeurWP}, $\modelfunc$ is defined as an encode-process-decode sequence using \glspl{GNN} \cite{keisler, graphcast}.
See \citeauth{gnn_intro} for an accessible introduction to \glspl{GNN}.
Input states are \textit{encoded} to the nodes of a \textit{mesh graph}, \textit{processed} through multiple \gls{GNN} layers and then \textit{decoded} back to the grid nodes to produce a prediction $\pred^{\timei}$.
The process is illustrated in \cref{fig:model_overview}.
The mesh graph contains fewer nodes than the grid, making it more efficient to work on.

In previous works, \citeauth{keisler} constructed the mesh graph as an icosahedral mesh of triangles around the globe. 
\citeauth{graphcast} extended this idea in the \gc model, by combining such icosahedral meshes at 7 resolutions.
These 7 graphs are merged by taking the union of their edge sets, creating a \textit{multi-scale} mesh graph with edges of varying length.
The idea behind the multi-scale graph is that longer edges can propagate information over longer spatial distances, which should be useful for forecasting physical processes taking place at larger spatial scales \cite{graphcast}.

We base our first model, titled \ms, on directly adapting the \gc model to the limited area setting.
This model produces one step predictions using the last two states ($\arorder=2$) \cite{graphcast}.
The previous weather states and known forcing inputs are encoded by \glspl{MLP} to vector representations in each grid node.
Additional static features, associated with edges and mesh nodes, are encoded using separate \glspl{MLP}.
As in \gc, all \glspl{GNN} used in each step of the encode-process-decode sequence are interaction networks \cite{interaction_nets}.
These networks update all node representations $\nodevec{\cdot}$ and edge representations $\edgevec{}{\cdot}{\cdot}$ according to
\alse[interaction_net]{
    \edgevecprime{}{s}{r} &\leftarrow \mlp{}{E}{\left[\edgevec{}{s}{r}, \nodevec{s}, \nodevec{r} \right]}\\
    \edgevec{}{s}{r} &\leftarrow \edgevec{}{s}{r} + \edgevecprime{}{s}{r},
    \qquad
    \nodevec{r} \leftarrow \nodevec{r} + \mlp{}{V}{\textstyle\left[\nodevec{r}, \sum_{s: (s,r) \in \edgeset{}} \edgevecprime{}{s}{r}\right]}
}
where $\edgeset{}$ is the set of edges used by the the current \gls{GNN}.
In the encoding step these edges go from grid nodes to mesh nodes, in the processing step they connect only mesh nodes and in the decoding step they go from the mesh back to the grid.
See \cref{sec:model_details} for model details.

\begin{figure}[tb]
    \centering
    \begin{subfigure}[b]{0.47\linewidth}
         \centering
         \includegraphics[width=\linewidth]{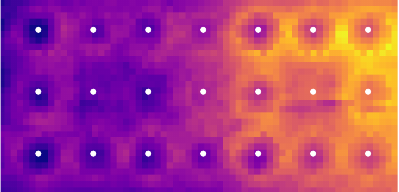}
         \caption{\ms}
         \label{fig:nlwrs_gt}
     \end{subfigure}%
     \hspace{0.06\textwidth}%
     \begin{subfigure}[b]{0.47\textwidth}
         \centering
         \includegraphics[width=\linewidth]{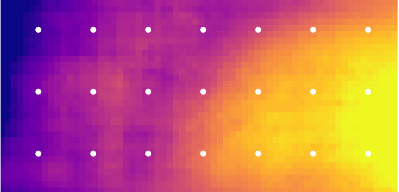}
         \caption{\hi}
         \label{fig:nlwrs_hi}
     \end{subfigure}
    \caption{
    Enlarged prediction of \wvar{\detokenize{z_1000}}.
    The \ms model shows circular artifacts centered at mesh nodes with $> 8$ neighbors (white dots).}
    \label{fig:artefact_vis}
\end{figure}

\paragraph{Hierarchical Graph Neural Network}
From initial experiments with the \ms model, we noticed issues with circular artifacts in the predictions.
The positions of these artifacts, which can be seen in \cref{fig:artefact_vis}, correspond to mesh nodes with many incoming edges in the multi-scale graph.
Motivated by this, we introduce the extended \hi model with a \textit{hierarchical} mesh graph in $\himaxl$ levels.
This hierarchy is created by not merging the $\himaxl$ meshes of different resolution levels used in the multi-scale graph, but instead introducing additional edges between the levels.
With such a hierarchical construction, illustrated in \cref{fig:model_overview}, we retain longer spatial edges at higher levels in the hierarchy, but get a more uniform structure in the bottom mesh level that is connected to the grid.

In the hierarchical model we extend the encoding step to not just propagate information from the grid to the bottom mesh level, but also up through the mesh hierarchy.
During the processing step we sequentially update the representations at each level in the hierarchy.
At each level we run one iterations of intra-level processing and then propagates information on to the next level, in the sequence $\himaxl,\dots, 1, \dots,\himaxl$.
It is also possible to run multiple such sweeps down and up through the hierarchy.
The decoding step then propagates information down through mesh levels $\himaxl, \dots, 1$ and finally to the grid for prediction.
All steps are parametrized using independent \glspl{GNN}.

\paragraph{Forecasting for a Limited Area}
A major part of adapting the graph-based models to the limited area setting is to redesign the mesh graph.
We will here specifically consider the operational area of the \gls{MEPS} forecasting system, but the method is general.
The MEPS area is defined by drawing a rectangular grid on a Lambert conformal conic map projection \cite{arome_metcoop}, which yields approximately equally sized grid cells.
We define our mesh graphs as regular quadrilateral meshes covering the same area, but with far less nodes than the original grid.
Regular meshes are created at $L=4$ different resolutions and either merged for \ms or connected with additional edges in \hi.
More details on the graph construction are given in \cref{sec:graph_construction}.

In limited area \gls{NWP}, the regional model needs information from the surrounding area. 
This is provided by updating lateral boundary conditions with data from a (often global) host model \cite{arome_metcoop, atmospheric_modeling, messinger2013}.
We follow the same methodology for our \gls{NeurWP} model, including \textit{boundary forcing} at each time step.
To incorporate this information, we let the model output an initial prediction $\pred^t$ and then replace each row $\pred^{t}_v$ with $\wstate^{t}_v$ if grid node $v$ lies within the boundary area.
When unrolling for multiple time steps this boundary forcing will then be fed back into the model through $\pred^{t}$.
Note that we here use the ground truth weather state in the dataset as the boundary forcing, but in operational scenarios this could be replaced by a forecast from an external host model.

\section{Experiments with MEPS Data}
\newcommand{\exppredfig}[2]{%
\begin{subfigure}[b]{0.24\textwidth}
     \centering
     \includegraphics[width=\textwidth]{graphics/nlwrs_pred/nlwrs_#1.pdf}
     \caption{#2}
\end{subfigure}}
\begin{figure}[tb]
    \centering
    \exppredfig{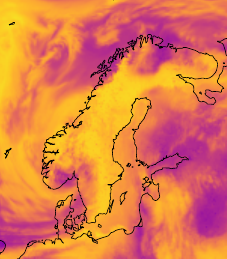}{Ground Truth}%
    \hspace{0.01\textwidth}%
    \exppredfig{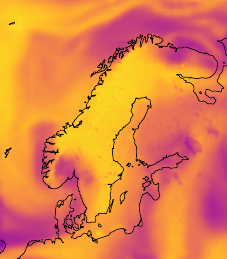}{\hi}%
    \hspace{0.01\textwidth}%
    \exppredfig{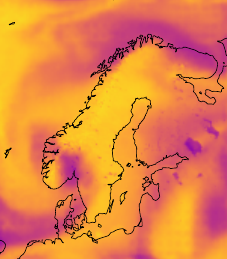}{\ms}%
    \hspace{0.01\textwidth}%
    \exppredfig{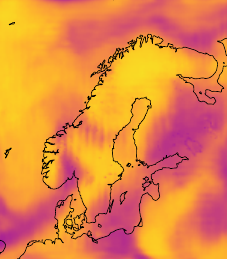}{\ol}%
    \caption{Ground truth and example forecasts of \wvar{nlwrs} at lead time 57 \si{\hour}.}
    \label{fig:example_forecasts}
\end{figure}
\begin{figure}[tb]
    \centering
    \begin{subfigure}[b]{0.49\textwidth}
        \centering
        \includegraphics[width=\textwidth]{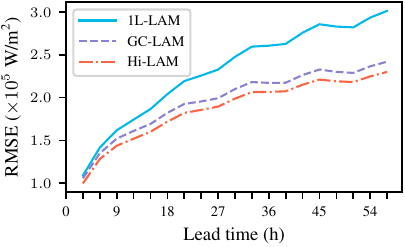}
        \caption{\wvar{nlwrs}}
        \label{fig:nlwrs_error}
    \end{subfigure}%
    \hspace{0.019\textwidth}%
    \begin{subfigure}[b]{0.49\textwidth}
        \centering
        \includegraphics[width=\textwidth]{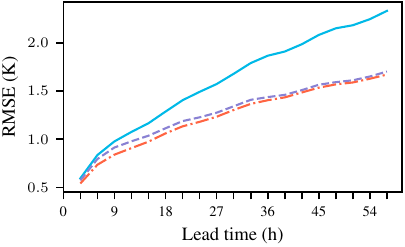}
        \caption{\wvar{\detokenize{t_2}}}
        \label{fig:t_2_error}
    \end{subfigure}
    \caption{\acrshort{RMSE} of net longwave solar radiation flux~(\wvar{nlwrs}) and temperature 2 \si{\metre} above ground~(\wvar{\detokenize{t_2}}), evaluated at different lead times.}
    \label{fig:experiment_line_plots}
\end{figure}

\paragraph{Dataset and Models}
We experiment with limited area \gls{NeurWP} on a dataset of historical forecasts from the \gls{MEPS} system \cite{arome_metcoop}.
Training on such data allows for building a fast surrogate model approximating the system.
The dataset contains 6069 forecasts from the years 2021--2023.
We use the first 15 months for training and validation and the remaining as a test set.
In total we model $S=17$ atmospheric variables, including temperature, wind and net solar radiation, among others (see details in \cref{sec:data_details}).
Our \gls{MEPS} data contains a grid of $238 \times 268$ nodes, resulting in a spatial resolution of \qty{10}{\kilo\metre}.
We evaluate the models based on \qty{57}{\hour} forecasts with \qty{3}{\hour} time steps.
This is close to the \qty{66}{\hour} forecasts used in the operational system \cite{arome_metcoop}.
Apart from the \ms and \hi models we also train a model (titled \ol) without any multi-scale edges, using a regular mesh of only a single resolution.
All models were trained first for single step prediction and then fine-tuned using 4 time step rollouts.

\paragraph{Results}
We showcase example forecasts in \cref{fig:example_forecasts} and \gls{RMSE} values for net longwave radiation (\wvar{nlwrs}) and 2 \si{\metre} temperature (\wvar{\detokenize{t_2}}) in \cref{fig:experiment_line_plots}.
More results, including errors and example forecasts for all variables can be found in \cref{sec:additional_res}.
While the \gls{RMSE} metric gives a useful overview of the model performance, equally important are the qualitative comparisons in \cref{fig:artefact_vis,fig:example_forecasts} and \cref{sec:example_predictions}.
In general we observe that the \hi model does not just reduce visual artifacts, but also gives overall lower \gls{RMSE}.
The performance of the \ol model is substantially worse than the others, verifying the importance of including edges of multiple spatial scales in the model.
The predictions of the \ol model also show severe visual artifacts.
Producing a 57 h forecast with \hi takes only \qty{1.5}{seconds} on a single A100 GPU, making the method promising for the type of large ensemble forecasting necessary to model extreme weather events. 

\paragraph{Future Work}
A general issue in existing \gls{NeurWP} models, ours included, is that they produce over-smooth, unrealistic forecasts \cite{weatherbench2, graphcast}. 
This is a consequence of the loss-functions used, making the models output only the mean of the predictive distribution.
To combat over-smoothing and increase the usefulness of \gls{NeurWP} models we believe that future works should move in the direction of more probabilistic and generative modeling.
For improving \gls{NeurWP} \glspl{LAM} there is also much to explore in how to best make use of learned representations or forecasts from existing global models.
In exemplifying a framework of graph-based \gls{NeurWP} \glspl{LAM} we hope to inspire the community to further explore also this modeling direction and its unique challenges.

\begin{ack}
We would like to thank Filip Ekström Kelvinius for helpful discussions about the \gls{GNN} implementation.
This research is financially supported by the Swedish Research Council via the project
\emph{Handling Uncertainty in Machine Learning Systems} (contract number: 2020-04122),
the Wallenberg AI, Autonomous Systems and Software Program (WASP) funded by the Knut and Alice Wallenberg Foundation,
the Excellence Center at Linköping--Lund in Information Technology (ELLIIT),
and
European Union's Horizon 2020 research and innovation program under grant agreements 646039, 775970 and 883973.
Our computations were enabled by the Berzelius resource at the National Supercomputer Centre, provided by the Knut and Alice Wallenberg Foundation.
\end{ack}

\small
\bibliography{references}
\normalsize

\newpage
\appendix
\section{Climate Impact}
\label{sec:climate_impact_app}
Accurate weather forecasting has always been massively useful, but in the time of climate change its importance is greater than ever.
The capabilities of \gls{NeurWP} systems introduce great new possibilities which can impact our ability to tackle climate change.
In many of these scenarios limited area \gls{NeurWP} can offer functionality that is of particular usefulness.

\paragraph{Extreme Weather}
Due to climate change the prevalence and severity of extreme weather events is expected to increase substantially, endangering both property and human life \cite{ipcc_2023}. 
These events are also getting harder to predict and the cost of damages per event has increased nearly 77\% over the past five decades \citeapp{world_economic_forum}.
In order to detect extreme events, there is a need to model the full distribution of possible weather states.
To achieve this, \textit{ensemble forecasts} are typically used, where multiple weather forecasts are produced from perturbed versions of initial states \cite{fundamentals_of_nwp}.
While ensemble forecasting historically has been limited by the computational costs of \gls{NWP} systems, efficient \gls{NeurWP} models has the potential to vastly improve our ability to model extreme weather.
For tracking extreme weather it is desirable to model at a high resolution mainly the area where the weather event is expected to unfold.
Such investigations are enabled by limited area \gls{NeurWP}.

\paragraph{Renewable Energy}
The production of renewable energy, such as solar and wind power, can be highly volatile \citeapp{re_integration}.
This creates challenges for including these sources in the larger energy system.
Accurate weather forecasts, translated into forecasts of energy production, fill an important role as enablers of these energy sources by making their output predictable.
Here, fast limited area \gls{NeurWP} models enable high resolution modeling in the areas where renewable energy is produced.
Detailed probability estimates from ensemble forecasting additionally allow for improved cost-loss decision making in these systems.

\paragraph{The Energy Footprint of \gls{NWP}}
Traditional \gls{NWP} systems utilize massive computing clusters \cite{fundamentals_of_nwp}, resulting in a substantial energy footprint of the forecasting process.
In comparison, \gls{NeurWP} models are highly energy efficient, even when taking into account their initial training process. 
The total energy required to train the large global \gls{NeurWP} model FourCastNet is comparable to running a 10 day forecast with 50 ensemble members using a traditional \gls{NWP} system \cite{fourcastnet}. 
Producing a forecast using the same model uses four orders of magnitude less energy than traditional \gls{NWP}.
A similar reduction in energy footprint is to be expected for \glspl{LAM}.

\paragraph{Interactive Digital Models}
By creating interactive digital models of the earth scientists and non-experts can perform simulations to aid policy-making and societal adaptation to climate change.
An example of such a model is the \textit{Destination Earth} initiative of the European Commission \citeapp{destination_earth}.
Fast and efficient \gls{NeurWP} is an important enabler for much of the monitoring and prediction capabilities of these models.
One promise of digital twin models is to be able to zoom in and with high-resolution details study extreme weather events.
Limited area \gls{NeurWP} is a natural and appealing choice to achieve such functionality.

\paragraph{National and Local Community Agency}
In a future where climate modeling is of increasing importance, both global and local models have important roles to play.
Global models, often enabled by international collaborative efforts, are important for understanding large-scale climate processes and provide necessary boundary information to \glspl{LAM} \cite{fundamentals_of_nwp, arome_metcoop}.
Local models complement these in allowing for studying regions of specific interest to single nations or communities.
Limited area \gls{NeurWP} bring the capabilities of machine learning models to local actors.
This allows them to use these models to forecast and study what is important to the local community they serve.
We hope that the effect of this will be increased agency of non-global organization to steer their own climate change adaptation efforts.

\section{Model Details}
\label{sec:model_details}
In this appendix we provide more details on the models used.

\subsection[GraphCast (GC-LAM and 1L-LAM)]{\gc{} (\ms{} and \ol{})} %
We start by re-iterating the steps of the \gc model from \citeauth{graphcast}.
\ms and \ol follow exactly the same procedure for making predictions, but using graphs adapted to our setting.
To compute a one time step prediction, the function $\modelfunc$ in \gc operates in three steps:
\newcommand{\procstep}[2]{\item \textbf{#1}: #2}
\begin{enumerate}
    \procstep{Encoding}{Encode grid node information to the mesh.}
    \procstep{Processing}{Perform multiple steps of processing on the mesh.}
    \procstep{Decoding}{Decode from the mesh nodes down to the grid nodes for prediction.}
\end{enumerate}
Each of the steps above is associated with one set of edges.
Grid2Mesh edges $\gtmedges$ are used during encoding, Mesh2Mesh edges $\mtmedges$ during processing and Mesh2Grid edges $\mtgedges$ during decoding.
The edges in $\gtmedges$ and $\mtgedges$ are directed, whereas $\mtmedges$ contains bidirectional edges\footnote{i.e.\ $\mtmedges$ also contains directed edges, but satisfies $(a,b) \in \mtmedges \Rightarrow (b,a) \in \mtmedges$.}.

\paragraph{Encoding}
The input to this whole procedure is a set of vector representations for nodes and edges.
Previous weather states, forcing inputs and static grid node features are concatenated and encoded using an \gls{MLP} into a vector $\gridnodevec{s}$ for each grid node $s$.
Similarly, static features of each mesh node $r$ are encoded into a vector representation $\meshnodevec{r}$.
Also the static features of edges are encoded into representations $\gtmvec{s}{r}$, $\mtmvec{s}{r}$ and $\mtgvec{s}{r}$ using separate \glspl{MLP}.
The notation $s \rightarrow r$ describes that a vector is associated with the edge from node $s$ to node $r$.
Each of the nodes $s$ and $r$ can belong either to the grid or the mesh depending on the context.
For example, in $\gtmvec{s}{r}$ the node $s$ belongs to the grid and $r$ to the mesh.
All \glspl{MLP} used have one hidden layer with Swish activation function \citeapp{swish} and are followed by a LayerNorm \citeapp{layernorm} layer.

The \gls{GNN} layers used in \gc are Interaction Networks \citep{interaction_nets}, which update node and edge representations using two \glspl{MLP}.
During encoding, one round of \gls{GNN} message passing is performed from the grid to the mesh.
The complete update being performed by this step is
\alse[gc_encoder_step]{
    \gtmvecprime{s}{r} &\leftarrow \gtmmlp{E}{\left[\gtmvec{s}{r}, \gridnodevec{s}, \meshnodevec{r} \right]}\\
    \gtmvec{s}{r} &\leftarrow \gtmvec{s}{r} + \gtmvecprime{s}{r}\label{eq:res_eq_start}\\
    \gridnodevec{s} &\leftarrow \gridnodevec{s} + \gtmmlp{G}{\gridnodevec{s}}\\
    \meshnodevec{r} &\leftarrow \meshnodevec{r} + \gtmmlp{V}{\left[\meshnodevec{r}, \sum_{s: (s,r) \in \gtmedges} \gtmvecprime{s}{r}\right]}\label{eq:res_eq_stop}
}
where $[\cdot]$ denotes vector concatenation.
The updates in each such step are always performed concurrently for all nodes or edges.
Note that \crefrange{eq:res_eq_start}{eq:res_eq_stop} describe residual connections in the updates of all node and edge representations.
The point of updating also the grid node representations here is that they will be used again in the decoder step.

\paragraph{Processing}
\label{sec:gc_processing}
The processing \glspl{GNN} operate only on the mesh nodes using $\mtmedges$.
There can be arbitrary many layers of processing \glspl{GNN} and these generally do not share parameters.
One processing layer $\playerindex$ performs the update
\alse[gc_processing_step]{
    \mtmvecprime{s}{r} &\leftarrow \mtmmlp{E, \playerindex}{\left[\mtmvec{s}{r}, \meshnodevec{s}, \meshnodevec{r} \right]}\\
    \mtmvec{s}{r} &\leftarrow \mtmvec{s}{r} + \mtmvecprime{s}{r}\\
    \meshnodevec{r} &\leftarrow \meshnodevec{r} + \mtmmlp{V, \playerindex}{\left[\meshnodevec{r}, \sum_{s: (s,r) \in \mtmedges} \mtmvecprime{s}{r}\right]}.
}

\paragraph{Decoding}
The decoder utilizes $\mtgedges$ to arrive at the final grid node representations as
\alse[gc_decoder_step]{
    \mtgvecprime{s}{r} &\leftarrow \mtgmlp{E}{\left[\mtgvec{s}{r}, \meshnodevec{s}, \gridnodevec{r} \right]}\\
    \gridnodevec{r} &\leftarrow \gridnodevec{r} + \mtgmlp{V}{\left[\gridnodevec{r}, \sum_{s: (s,r) \in \mtgedges} \mtgvecprime{s}{r}\right]}.
}
Note that there is no need to update the edge or mesh node representations at this point.
The final predicted weather state in each grid node $r$ is then computed as
\eq[gc_prediction]{
    \pred_{r}^\timei = \modelfunc\left(\wstate^{\timei-1}, \wstate^{\timei-2}\right)_r = \wstate^{\timei-1}_r + \mlp{\text{pred}}{}{\gridnodevec{r}}.
}
Note that \cref{eq:gc_prediction} includes a residual connection to the previous weather state, meaning that the output of the model (from $\mlpfunc{\text{pred}}{}$) is the state difference rather than the new state.
No LayerNorm is used in $\mlpfunc{\text{pred}}{}$.

\subsection{\hi}
\paragraph{Motivation}
\gc augments the original graph construction from \citeauth{keisler} by the addition of multi-scale edges, but does not introduce extra nodes \cite{graphcast}.
When the meshes of multiple resolutions are merged, the final multi-scale graph will share its node set with the finest resolution mesh.
Nodes with long incoming multi-scale edges will aggregate information from longer spatial distances.
This information can then spread locally to neighboring nodes through additional layers of processing \glspl{GNN}.
However, since no new nodes are introduced in the multi-scale architecture, some nodes will 
\begin{inparaenum}[1)]
    \item have many more neighbors than others 
    \item aggregate both highly local and long-range information.
\end{inparaenum}
All nodes need to contain relevant local information, which should be decoded to the connected grid nodes for prediction.
Nodes with long multi-scale edges will thus have a dual-responsibility: retain local information and spread long-range information. 
We hypothesize that these nodes with more neighbors will see a distributional shift in their representation vectors, which would explain the artifacts observed in \cref{fig:artefact_vis}.
As the \gls{GNN} parameters in each layer are shared for all mesh nodes these issues could make it hard for the model to learn useful processing layers.

We propose to not just add multi-scale edges, but create a hierarchical structure of nodes and edges that alleviates the problems mentioned above.
To achieve this, we extend the encoding and decoding steps of \gc and fully redefine the processing step.
\paragraph{Hierarchical Structure}
Our hierarchical multi-scale mesh graph consists of $L$ levels of nodes, with level 1 at the bottom (connected to the grid) and level $L$ at the top.
Higher levels in the hierarchy contain less nodes with longer spatial distances between them.
While we think of and visualize these levels at different height there is no vertical component to the node positions and all graphs lay flat on the surface of the globe.
Let $\levelnodevec{l}{r}$ denote the vector representation of mesh node $r$ in level $l$.
We retain the edge sets $\gtmedges$ and $\mtgedges$ from earlier, but connect grid nodes only to mesh nodes at the lowest level $l=1$.
Instead of a single set of mesh edges, the hierarchical setup will require multiple edge sets connecting nodes both inside and between levels.
For each mesh level $l$ we define $\msameedges{l}$ as the (bidirectional) edges connecting nodes internally in this level.
To propagate information up in the hierarchy, we define sets $\mupedges{l}{l+1}$ for $l \in \set{1, \dots, L-1}$ with directed edges from nodes in level $l$ to nodes in level $l+1$.
Similarly, we define downward edges $\mdownedges{l}{l-1}$ for $l \in \set{2, \dots, L}$.
Static features for all these edges are initially encoded into representations $\msamevec{l}{s}{r}$, $\mupvec{l}{l+1}{s}{r}$, $\mdownvec{l}{l-1}{s}{r}$.

\paragraph{Encoding}
The encoding in \cref{eq:gc_encoder_step} will only spread information from the grid nodes to the bottom level of the mesh hierarchy.
If we would start the processing from there, the upper levels of the hierarchy would initially only be processing information from static features.
As these static features are generally not key to making the prediction, this would be a waste of computation and memory.

In order to avoid this we augment the encoding step to propagate information up from the grid to the top of the hierarchy.
The new encoding step starts by encoding from the grid to the bottom mesh level as
\alse[hi_encoder_step_g2m]{
    \gtmvecprime{s}{r} &\leftarrow \gtmmlp{E}{\left[\gtmvec{s}{r}, \gridnodevec{s}, \levelnodevec{1}{r} \right]}\\
    \gtmvec{s}{r} &\leftarrow \gtmvec{s}{r} + \gtmvecprime{s}{r}\\
    \gridnodevec{s} &\leftarrow \gridnodevec{s} + \gtmmlp{G}{\gridnodevec{s}}\\
    \levelnodevec{1}{r} &\leftarrow \levelnodevec{1}{r} + \initnodemlp{1}{\left[
        \levelnodevec{1}{r}, 
        \smashoperator[r]{\sum_{s: (s,r) \in \gtmedges}}
            \gtmvecprime{s}{r}
    \right]}
}
and then continues up through the hierarchy
\alse[hi_encoder_step_up]{
    \mupvecprime{l-1}{l}{s}{r} &\leftarrow \initedgemlp{l-1}{l}{\left[\mupvec{l-1}{l}{s}{r}, \levelnodevec{l-1}{s}, \levelnodevec{l}{r}\right]}\\
    \mupvec{l-1}{l}{s}{r} &\leftarrow \mupvec{l-1}{l}{s}{r} + \mupvecprime{l-1}{l}{s}{r}\\
    \levelnodevec{l}{r} &\leftarrow \levelnodevec{l}{r} + \initnodemlp{l}{\left[
        \levelnodevec{l}{r}, 
        \smashoperator[r]{\sum_{s: (s,r) \in \mupedges{l-1}{l}}}
            \mupvecprime{l-1}{l}{s}{r}
    \right]}
}
which is ran sequentially for level $l=2 \dots, \himaxl$.
\Cref{eq:hi_encoder_step_up} propagates information up from the previous level through $\levelnodevec{l-1}{s}$ and combines this with the static feature embeddings.

\paragraph{Processing}
We redefine the processing step to sequentially pass through the different levels of the hierarchy, using independent \glspl{MLP} at each level.
One processing layer $\playerindex$ performs a sweep from the top level $L$, down to the bottom mesh level 1 and the all the way up to level $L$ again.
The steps of this process alternate between intra-level processing at each level and message passing to the next level in the sweep.
In more detail, the downward part of the process is described by
\alse[hi_process_down]{
    \msamevecprime{l}{s}{r} &\leftarrow \msamemlp{l}{\playerindex\downarrow}{\left[\msamevec{l}{s}{r}, \levelnodevec{l}{s}, \levelnodevec{l}{r}\right]}\\
    \msamevec{l}{s}{r} &\leftarrow \msamevec{l}{s}{r} + \msamevecprime{l}{s}{r}\\
    \levelnodevec{l}{r} &\leftarrow \levelnodevec{l}{r} + \mupdatemlp{l}{\playerindex\downarrow}{\left[\levelnodevec{l}{r}, \smashoperator[r]{\sum_{s: (s,r) \in \msameedges{l}}} \msamevecprime{l}{s}{r} \right]}\\
    \mdownvecprime{l}{l-1}{s}{r} &\leftarrow \mdownmlp{l}{l-1}{\playerindex}{\left[\mdownvec{l}{l-1}{s}{r}, \levelnodevec{l}{s}, \levelnodevec{l-1}{r}\right]}\label{eq:hi_down_skip_start}\\
    \mdownvec{l}{l-1}{s}{r} &\leftarrow \mdownvec{l}{l-1}{s}{r} + \mdownvecprime{l}{l-1}{s}{r}\\
    \levelnodevec{l-1}{r} &\leftarrow \levelnodevec{l-1}{r} + \mupdatemlp{l-1}{\playerindex\searrow}{\left[\levelnodevec{l-1}{r}, \smashoperator[r]{\sum_{s: (s,r) \in \mdownedges{l}{l-1}}} \mdownvecprime{l}{l-1}{s}{r}\right]}\label{eq:hi_down_skip_end}
}
ran sequentially for levels $l=\himaxl,\dots, 1$ (skipping \crefrange{eq:hi_down_skip_start}{eq:hi_down_skip_end} on level $l=1$).
Then, the process continues back up as 
\alse[hi_process_up]{
    \msamevecprime{l}{s}{r} &\leftarrow \msamemlp{l}{\playerindex\uparrow}{\left[\msamevec{l}{s}{r}, \levelnodevec{l}{s}, \levelnodevec{l}{r}\right]}\\
    \msamevec{l}{s}{r} &\leftarrow \msamevec{l}{s}{r} + \msamevecprime{l}{s}{r}\\
    \levelnodevec{l}{r} &\leftarrow \levelnodevec{l}{r} + \mupdatemlp{l}{\playerindex\uparrow}{\left[\levelnodevec{l}{r}, \smashoperator[r]{\sum_{s: (s,r) \in \msameedges{l}}} \msamevecprime{l}{s}{r} \right]}\\
    \mupvecprime{l}{l+1}{s}{r} &\leftarrow \mupmlp{l}{l+1}{\playerindex}{\left[\mupvec{l}{l+1}{s}{r}, \levelnodevec{l}{s}, \levelnodevec{l+1}{r}\right]}\label{eq:hi_up_skip_start}\\
    \mupvec{l}{l+1}{s}{r} &\leftarrow \mupvec{l}{l+1}{s}{r} + \mupvecprime{l}{l+1}{s}{r}\\
    \levelnodevec{l+1}{r} &\leftarrow \levelnodevec{l+1}{r} + \mupdatemlp{l+1}{\playerindex\nearrow}{\left[\levelnodevec{l+1}{r},  \smashoperator[r]{\sum_{s: (s,r) \in \mupedges{l}{l+1}}} \mupvecprime{l}{l+1}{s}{r} \right]}\label{eq:hi_up_skip_end}
}
ran sequentially for levels $l=1,\dots,\himaxl$ (skipping \crefrange{eq:hi_up_skip_start}{eq:hi_up_skip_end} on level $l=L$).
It is possible to use multiple processing layers, meaning that multiple such sweeps throughout the hierarchy will be performed.
As each edge set and mesh level has its own \glspl{MLP}, we arrive at a more flexible parametrization than the \gc mesh processing.
In particular, we learn separate \glspl{MLP} for each spatial scale, corresponding to the different levels in our hierarchy.

\paragraph{Decoding}
Similarly to the encoding, we change the decoding step to sequentially aggregate information from the top of the mesh hierarchy down to the lowest level.
If we consider each level of the hierarchical mesh graph to model physical processes at a specific spatial scale, this type of decoding has a natural interpretation as aggregating all of this information for the final prediction.
We update the node representations according to
\alse[hi_decode]{
\mdownvecprime{l+1}{l}{s}{r} &\leftarrow \readoutedgemlp{l+1}{l}{\left[\mdownvec{l+1}{l}{s}{r}, \levelnodevec{l+1}{s}, \levelnodevec{l}{r}\right]}\\
\levelnodevec{l}{r} &\leftarrow \levelnodevec{l}{r} + \readoutnodemlp{l}{\left[
    \levelnodevec{l}{r}, 
    \smashoperator[r]{\sum_{s: (s,r) \in \mdownedges{l+1}{l}}}
        \mdownvecprime{l+1}{l}{s}{r}
\right]}
}
sequentially for level $l= \himaxl-1, \dots, 1$.
This is then followed by the final decoding to the grid
\alse[he_decode_grid]{
    \mtgvecprime{s}{r} &\leftarrow \mtgmlp{E}{\left[\mtgvec{s}{r}, \levelnodevec{1}{s}, \gridnodevec{r} \right]}\\
    \gridnodevec{r} &\leftarrow \gridnodevec{r} + \mtgmlp{V}{\left[\gridnodevec{r}, \sum_{s: (s,r) \in \mtgedges} \mtgvecprime{s}{r}\right]}.
}
Note that there is no need to update the edge representations during decoding, as they will not be used in any further processing.
The final prediction step is the same as in \gc (\cref{eq:gc_prediction}).

\subsection{Boundary Forcing}
\label{sec:boundary_forcing}
\begin{figure}
    \centering
    \includegraphics[width=\linewidth]{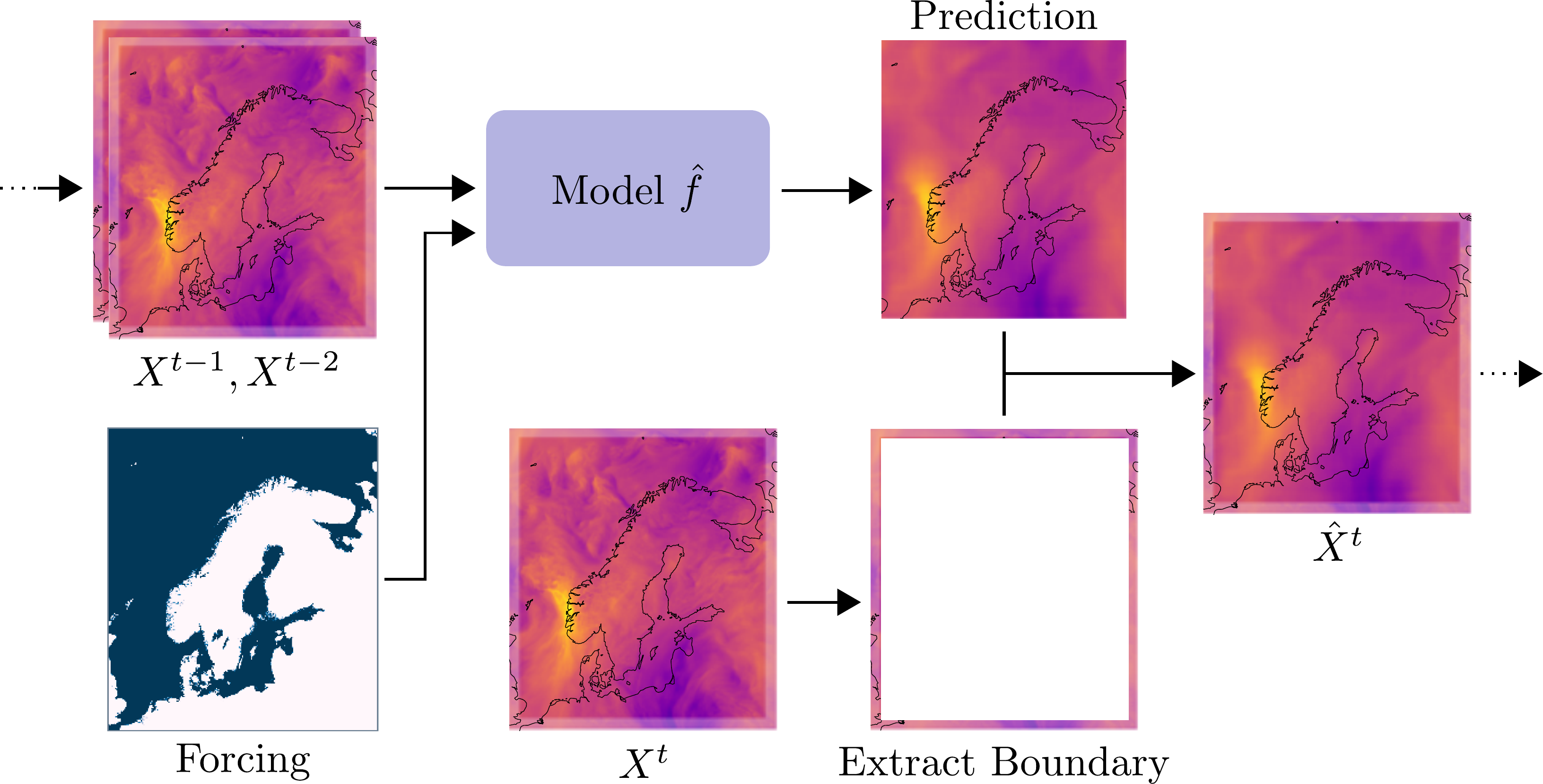}
    \caption{Schematic showing how boundary forcing is combined with the prediction during each autoregressive step.}
    \label{fig:boundary_forcing}
\end{figure}

In all models we use boundary forcing in order to include information about the surrounding area.
The boundary forcing is included at each time step by replacing predictions inside of the boundary area $\boundary$ with the ground truth forecast $\wstate^t$.
After making the prediction as in \cref{eq:gc_prediction}, we update the row for each node $v$ as
\eq[boundary_forcing_eq]{
    \pred^{t}_v \leftarrow \left(1 - \indicator{v \in \boundary}\right)\pred^{t}_v  + \indicator{v \in \boundary}\wstate^{t}_v
}
where $\indicator{\cdot}$ is the indicator function.
This process is also described in \cref{fig:boundary_forcing}.
Predictions for grid nodes within the boundary area are not used in the training loss or evaluation.

We define the boundary as the 10 closest grid nodes to the edge of the limited area.
Note that we use a boundary area that lays \textit{inside} the \gls{MEPS} area, allowing us to use the ground truth forecasts as boundary forcing.
In the operational system the boundary area is defined along the edge \textit{outside} the \gls{MEPS} area.
There is no major conceptual difference between these and one could easily re-define the different areas to match the operational \gls{MEPS} system.

\subsection{Loss Function}
All models are trained with a weighted \gls{MSE} loss similar to \gc.
We use batched training with the loss for each sample computed as
\al[loss_function]{
    \loss = 
    \frac{1}{\rolloutlen}\sum_{t=\inittime + 1}^{\inittime + \rolloutlen} 
    \frac{1}{\setsize{\interior}} \sum_{v \in \interior} 
    \sum_{i=1}^{\statedim}
    \invdiffweight_i \stateweight_i
    \left(\pred^t_{v,i} - \wstate^t_{v,i}\right)^2
}
where 
\begin{itemize}
    \item $\inittime$ is the initial time step the forecast starts from,
    \item $\rolloutlen$ is the number of rollout steps used during training,
    \item $\interior = \set{1, \dots, \ngridpoints} \setminus \boundary$ is the set of grid nodes not on the boundary,
    \item $\invdiffweight_i$ is the inverse variance of time differences for variable $i$,
    \item $\stateweight_i$ is a weight associated with the vertical level of variable $i$.
\end{itemize}
The sums in the loss are (in order) over
\begin{inparaenum}[1)]
\item forecast time steps,
\item grid nodes and 
\item weather state variables. 
\end{inparaenum}
We refer to \citeauth{graphcast} for details about the variable weighting scheme.
Note that as the \gls{MEPS} area has grid cells of approximately equal size we do not need to weight the grid nodes.

\subsection{Training Details}
Our trained models use latent representations of dimensionality 64 everywhere.
The \ms and \ol models use 4 processing layers, and for \hi we use only 2.
With the sweep down and up through the hierarchy each processing layer of \hi updates node and edge representations twice.
To keep a fair comparison we thus use half as many processing layers.
With these architectural choices we are able to train models with a batch size of 8 using a single NVIDIA A100 GPU with \qty{80}{\giga\byte} VRAM.
All models were trained for single-step prediction for 500 epochs and then fine-tuned using 4 time-step rollouts for 200 epochs.
The AdamW optimizer \citeapp{adamw} was used with a learning rate of 0.001.
Training each model takes 3--4 days in total.
The \hi model, despite being larger and more sequential in nature, only takes 12\% longer to train than \ms.
Preliminary experiments showed only minor improvements when fine-tuning on longer rollouts, even when considering the error at the highest lead times.

\section{Data Details}
\label{sec:data_details}
In this appendix we give more details on the exact features and dataset used in our experiments.
While some of the forcing and static features follow the exact format of \gc \cite{graphcast}, we here describe everything for the sake of completeness.

\subsection{Atmospheric Variables}
\newcommand{\varline}[5]{\wvar{\detokenize{#1}} & #2 & #3 & #4 & \parbox{0.30\textwidth}{#5}\\}
\newcommand{\mepsbottomlevel}{Lowest \gls{MEPS} level\footnotemark[1]}
\newcommand{\geopotheight}[1]{#1 \si{\hecto\pascal} pressure}
\begin{table}[tbp]
\centering
\caption{The 17 weather variables in our dataset. \textsuperscript{1}In the \gls{MEPS} system 65 vertical levels are defined from the ground to the top of the atmosphere \cite{arome_metcoop}. The lowest \gls{MEPS} level sits at approximately 12.5 \si{\meter} above ground.}
\label{tab:variables_table}
\begin{tabular}{@{}lllcl@{}}
\toprule
\textbf{Abbreviation} & \textbf{Quantity} & \textbf{Vertical Level} & \multicolumn{1}{c}{\textbf{Unit}} & \textbf{Description} \\ \midrule
    \varline{pres_0g}{Pressure}{Ground level}{\si{\pascal}}{Atmospheric pressure at ground level\vspace{.3em}} %
    \varline{pres_0s}{Pressure}{Sea level}{\si{\pascal}}{Atmospheric pressure at sea level}
    \midrule
    \varline{nlwrs}{Solar radiation}{Ground level}{\si{\watt\per\metre\squared}}{Net longwave radiation flux} %
    \varline{nswrs}{Solar radiation}{Ground level}{\si{\watt\per\metre\squared}}{Net shortwave radiation flux}
    \midrule
    \varline{r_2}{Humidity}{2 \si{\metre} above ground}{-}{Relative humidity, in $[0,1]$}
    \varline{r_65}{Humidity}{\mepsbottomlevel}{-}{Relative humidity, in $[0,1]$}
    \midrule
    \varline{t_2}{Temperature}{2 \si{\metre} above ground}{\si{\kelvin}}{Instantaneous temperature}
    \varline{t_65}{Temperature}{\mepsbottomlevel}{\si{\kelvin}}{Instantaneous temperature}
    \varline{t_500}{Temperature}{\geopotheight{500}}{\si{\kelvin}}{Instantaneous temperature}
    \varline{t_850}{Temperature}{\geopotheight{850}}{\si{\kelvin}}{Instantaneous temperature}
    \midrule
    \varline{u_65}{Wind}{\mepsbottomlevel}{\si{\meter\per\second}}{$u$-component of wind}
    \varline{v_65}{Wind}{\mepsbottomlevel}{\si{\meter\per\second}}{$v$-component of wind}
    \varline{u_850}{Wind}{\geopotheight{850}}{\si{\meter\per\second}}{$u$-component of wind}
    \varline{v_850}{Wind}{\geopotheight{850}}{\si{\meter\per\second}}{$v$-component of wind}
    \midrule
    \varline{wvint}{Water vapor}{All}{\si{\kilo\gram\per\meter\squared}}{Integrated column of water vapor above grid cell}
    \midrule
    \varline{z_500}{Geopotential}{\geopotheight{500}}{\si{\meter\squared\per\second\squared}}{Instantaneous geopotential}
    \varline{z_1000}{Geopotential}{\geopotheight{1000}}{\si{\meter\squared\per\second\squared}}{Instantaneous geopotential}
\bottomrule
\end{tabular}
\end{table}
At each grid cell we model 17 weather variables, including a broad range of different quantities and different vertical levels in the atmosphere.
All variables are described in \cref{tab:variables_table}.
The particular choice of variables was motivated by a combination of meteorological relevance, data availability and striving for a diverse set of variables to evaluate the model on. 
For solar radiation (\wvar{nlwrs} and \wvar{nswrs}) we consider the net flux at ground level, aggregated over the past \qty{3}{hours} (since the last time step).
Apart from the solar radiation all other quantities are instantaneous.
For training we rescale the values of each variable to zero mean and unit variance.

\subsection{Forcing}
As forcing at each time point we include at each grid node:
\begin{itemize}
    \item The solar radiation flux at the top of the atmosphere.
    \item $(\sin(2 \pi \timeprogress / \timelength) + 1)/2$ and $(\cos(2 \pi \timeprogress / \timelength) + 1)/2$, 
    where $\timeprogress$ is the time of day and $\timelength$ the length of one day.
    \item The same sine and cosine features as above, but with $\timeprogress$ as the time in the current year and $\timelength$ as the length of the year.
    \item The fraction of open water in the grid cell. 
    We assume this to be constant over the forecast period and take the value from the time of the initial state.
    In \gc the open water mask is assumed to be fully static.
    Treating this as forcing could be useful for taking into account seasonal fluctuations of the ice cover in the Nordic region.
\end{itemize}
As in \gc we include the forcing for time steps $t-1$, $t$ and $t+1$ as input at time step $t$, for making the prediction $\pred^{t+1}$.
The boundary forcing applied at each time step (as described in \cref{sec:boundary_forcing}) consists of the same 17 weather variables as listed in \cref{tab:variables_table}.

\subsection{Static Features}
\paragraph{Grid Nodes}
The static features used at each grid node are: 
\begin{itemize}
    \item The 2-dimensional coordinate of the node in the \gls{MEPS} Lambert projection, normalized by the maximum coordinate value.
    \item The surface geopotential (topography).
    \item A binary indicator describing if the node is in the boundary area or not.
\end{itemize}

\paragraph{Mesh Nodes}
The only static features associated with mesh nodes are their coordinates, following the same format as for the grid nodes.

\paragraph{Edges}
All edges use the same type of static features. 
This includes 
\begin{itemize}
    \item The length of the edge.
    \item The vector difference between the source and target nodes, using the \gls{MEPS} Lambert projection coordinates.
\end{itemize}
Both of these features are normalized by the length of the longest edge.

\subsection{Dataset Details}
\begin{figure}
    \centering
    \includegraphics[width=\linewidth]{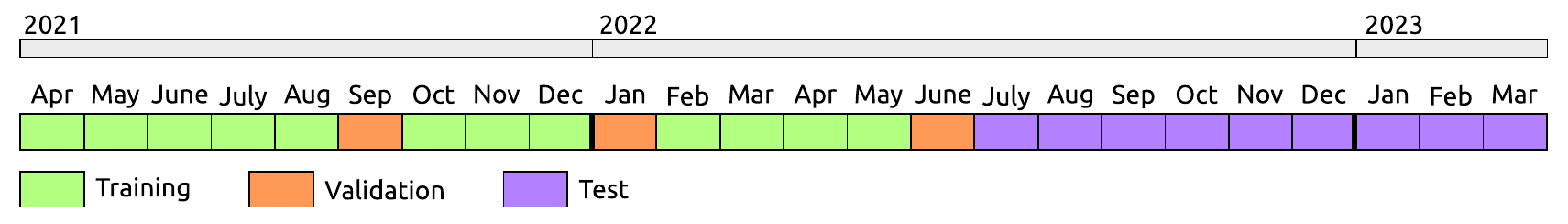}
    \caption{Overview of the period covered in the dataset and the training/validation/test split.}
    \label{fig:dataset_split}
\end{figure}

Our dataset consists of archived forecasts from the operational \gls{MEPS} system during the period April~2021 -- March~2023.
This period was chosen due to the system configuration being reasonably stable, preventing distributional shifts within the dataset.
From the chosen period we extract the forecasts started at 00 and 12 UTC each day.
At each starting time there are 5 ensemble forecasts, resulting in 10 forecasts per day (if no ensemble members are missing due to operational issues).
When retrieving the data we additionally downsample the spatial resolution from the original \qty{2.5}{\kilo\metre} to \qty{10}{\kilo\metre}.
This results in a dataset of 6069 forecasts of length \qty{66}{\hour} with \qty{1}{\hour} time steps.
We split these into training, validation and test sets according to \cref{fig:dataset_split}.
It should be noted that there is high correlation between many of these 6069 samples, in particular at early time steps in the forecasts.
The actual information content in the data, in terms of the effective sample size, is substantially lower.

The original data uses \qty{1}{\hour} time steps, but our \gls{NeurWP} models predict in \qty{3}{\hour} steps.
Because of this we can extract 3 training samples from each forecast in the dataset (i.e. original time steps $\set{1,4,7,\dots}$, $\set{2,5,8,\dots}$ and $\set{3,6,9,\dots}$).
As we train on only $\leq 4$ rollout steps, we also only need a subset of each such time series for each training iteration.
To make maximum use of the data during training we randomly sample which time step to use as initial state for unrolling the model.

The combination of 
\begin{inparaenum}[1)]
    \item the \qty{3}{\hour} time steps,
    \item using the last two weather states as model inputs,
    \item including forcing from multiple times as input,
\end{inparaenum}
means that we reduce the effective length of our ground truth forecasts in pre-processing.
This explains why we predict for \qty{57}{\hour} rather than the original \qty{66}{h}.
Note however that nothing prevents us from unrolling the models to create forecasts for \qty{66}{h} or beyond.
Although this is possible, we do not have ground truth data to compare against past \qty{57}{h} and therefore view such experiments to be of limited interest.
There is still no reason to expect the model performance to become drastically worse specifically past \qty{57}{h}, as all models are anyhow fine-tuned only on 4 time step rollouts (= \qty{12}{\hour} $\ll$ \qty{57}{\hour}).
\section{Graph Construction}
\label{sec:graph_construction}
We here give additional details on how the different graphs were constructed for the \gls{MEPS} area.

\subsection{Mesh Graph}
\begin{figure}[tbp]
    \begin{subfigure}[b]{0.5\textwidth}
        \centering
        \includegraphics[width=\textwidth]{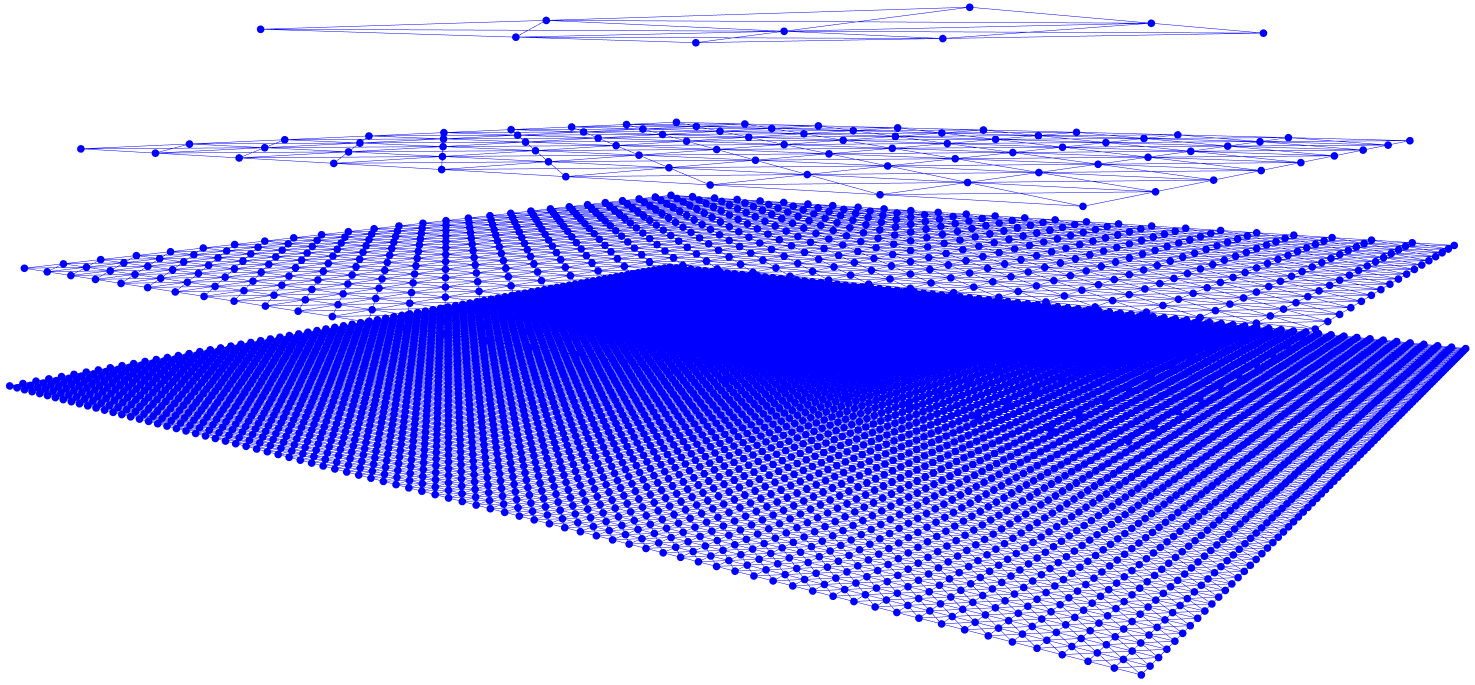}
        \caption{Initial mesh graphs at 4 resolutions}
        \label{fig:mesh_levels}
    \end{subfigure}%
    \begin{subfigure}[b]{0.5\textwidth}
        \centering
        \includegraphics[width=\textwidth]{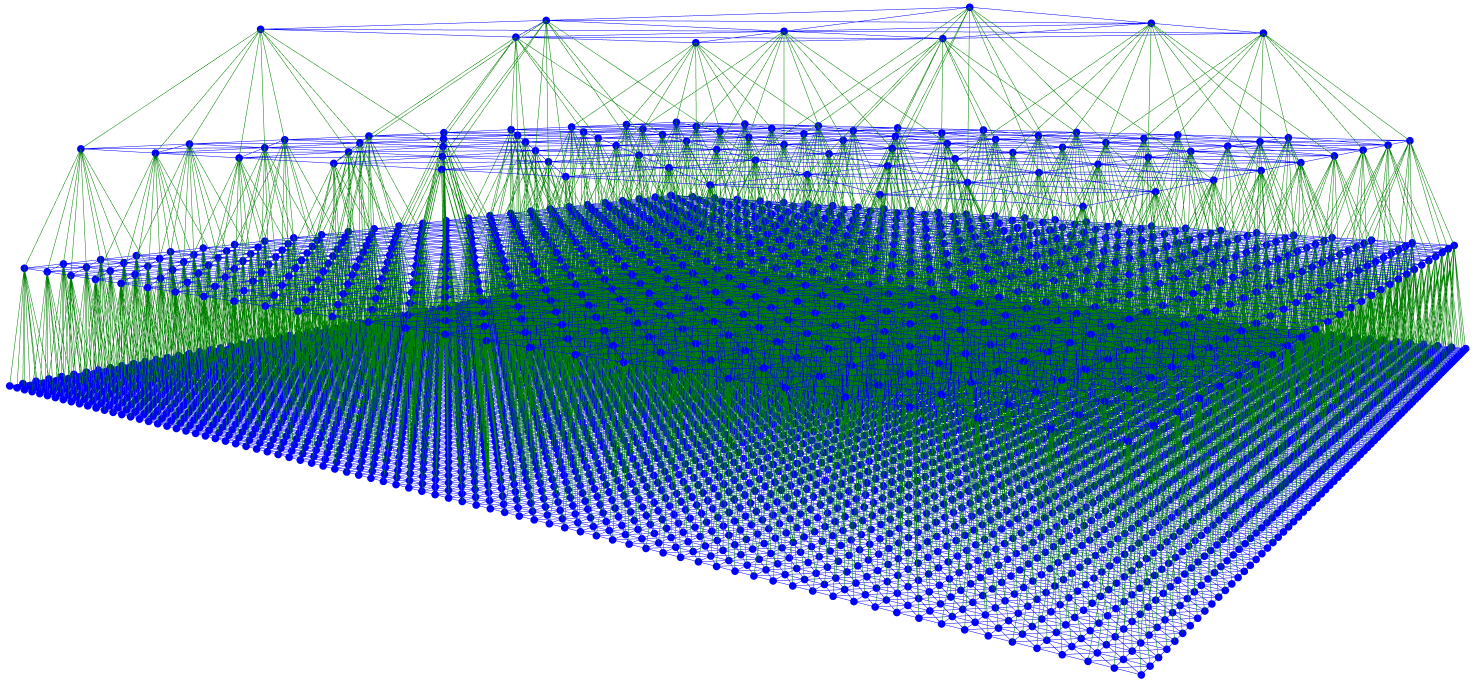}
        \caption{\hi mesh graph}
        \label{fig:hi_mesh}
    \end{subfigure}
    \begin{subfigure}[b]{0.5\textwidth}
        \centering
        \includegraphics[width=\textwidth]{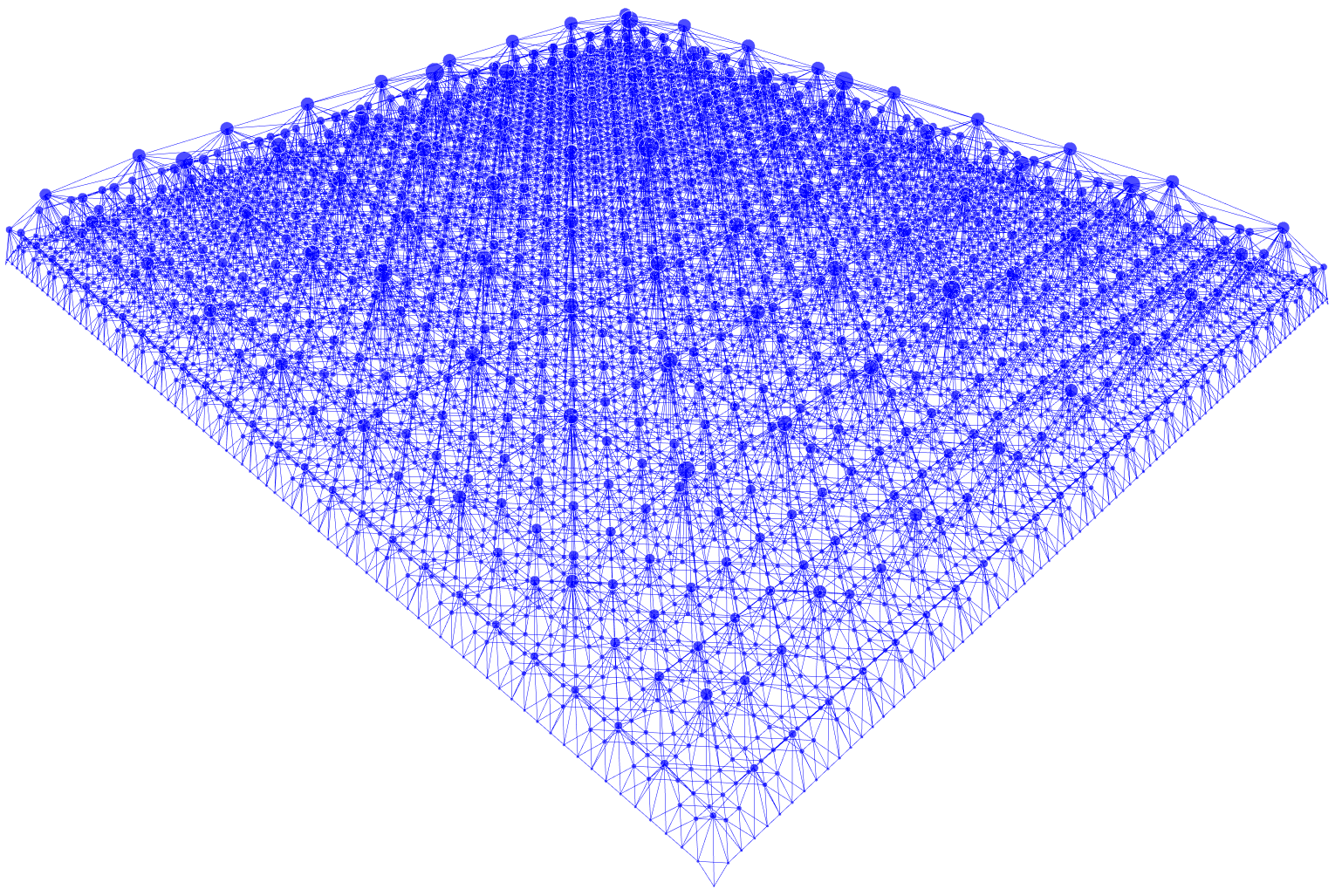}
        \caption{\ms mesh graph}
        \label{fig:ms_mesh}
    \end{subfigure}%
    \begin{subfigure}[b]{0.5\textwidth}
        \centering
        \includegraphics[width=\textwidth]{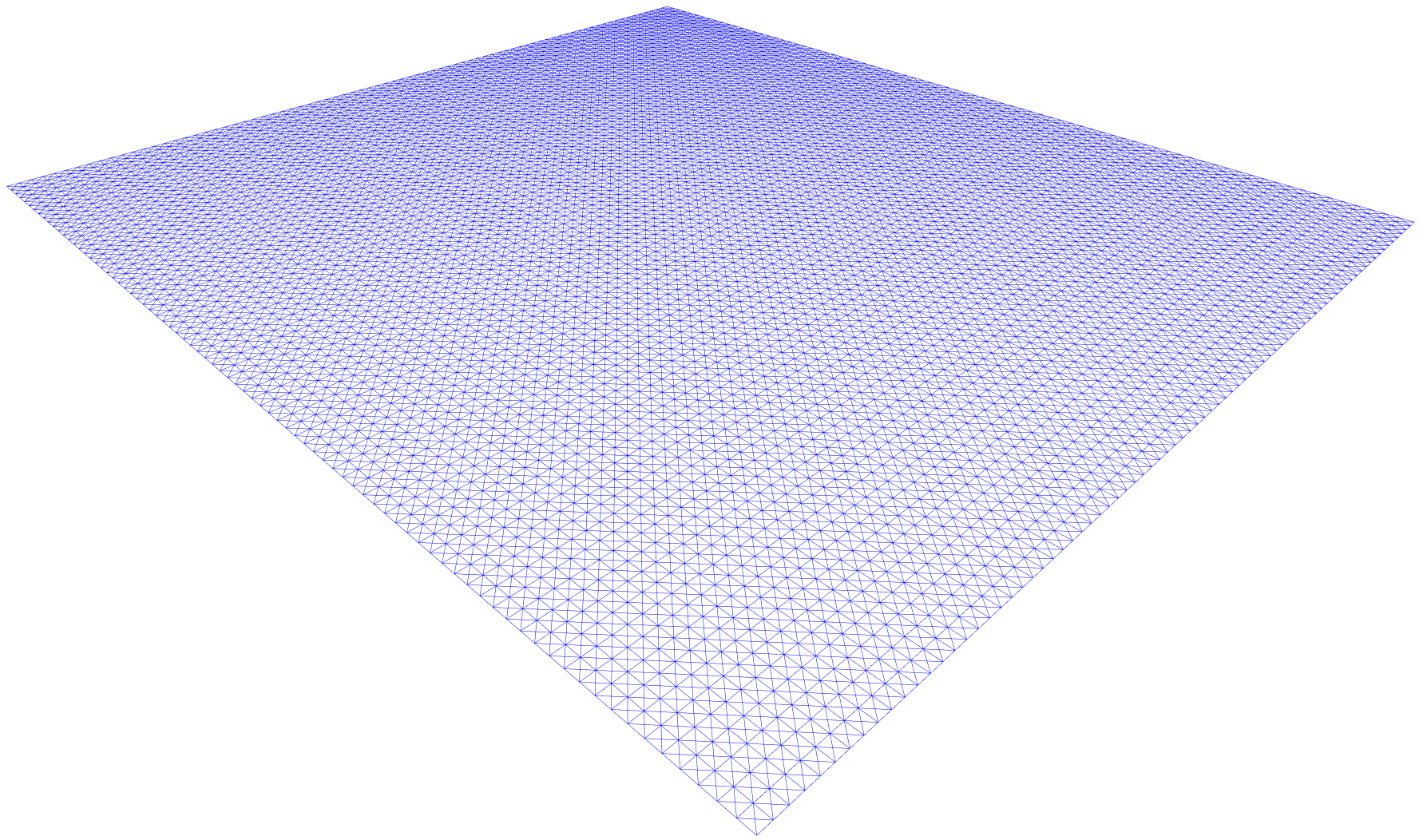}
        \caption{\ol mesh graph}
        \label{fig:ol_mesh}
    \end{subfigure}%
    \caption{Mesh graphs used in the different models. Note that the vertical positioning of nodes is purely for visualization purposes.
    In the \ms mesh graph the node size and vertical position is determined by its number of neighbors, to highlight the non-uniformity of the multi-scale structure.}
    \label{fig:3d_graphs}
\end{figure}

Our mesh graphs contain regular quadrilateral meshes covering the \gls{MEPS} area.
To create these we lay out mesh nodes in regular rows and columns over the area.
Each node is then connected with bidirectional edges to its neighbors horizontally, vertically and diagonally (see \cref{fig:graph_connections}).
This results in all nodes (except those at the edge of the area) having 8 neighbors.
The procedure is repeated at 4 different resolutions (\cref{fig:mesh_levels}), tripling the distance between nodes at each resolution.
This means that a node at resolution level $l$ is positioned at the center of a group of $3 \times 3$ nodes at resolution level $l-1$, sharing its exact position with the center node of the group (illustrated in \cref{fig:graph_alignment}).

\newcommand{\tikzring}[1]{\tikz\draw[black,radius=#1,very thick] (0,0) circle ;}%
\begin{figure}[tbp]
    \centering
    \begin{subfigure}[t]{0.35\textwidth}
        \centering
        \includegraphics[width=\textwidth]{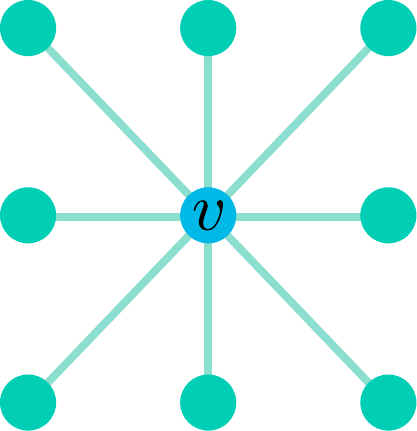}
        \caption{Each mesh node $v$ is connected to its neighbors horizontally, vertically and diagonally.}
        \label{fig:graph_connections}
    \end{subfigure}%
    \hspace{0.1\textwidth}
    \begin{subfigure}[t]{0.45\textwidth}
        \centering
        \includegraphics[width=\textwidth]{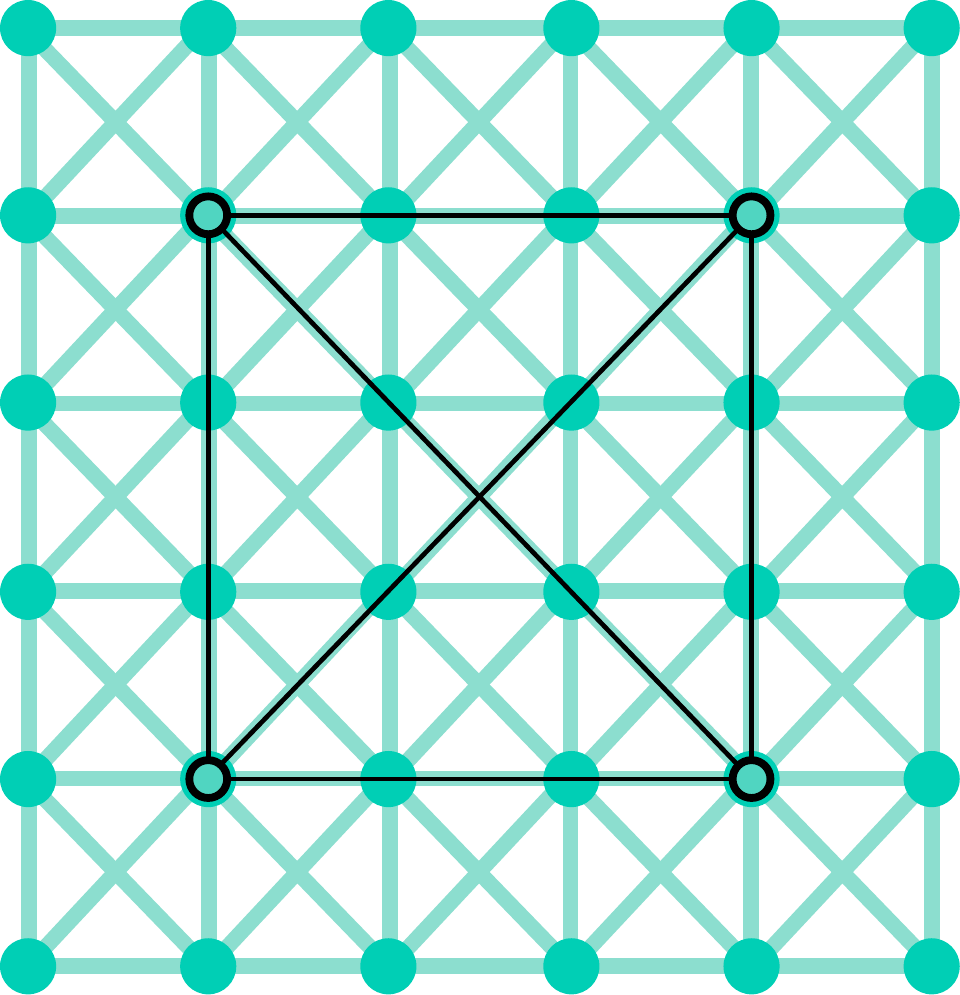}
        \caption{Alignment of mesh nodes \tikzring{2.5pt} at resolution level $l$ with mesh nodes \tikzcircle[LiUgreen,fill=LiUgreen]{3pt} at resolution level $l-1$.}
        \label{fig:graph_alignment}
    \end{subfigure}
    \caption{Illustration of the mesh graph construction process.}
    \label{fig:graph_construction}
\end{figure}

To create the multi-scale mesh graph for \ms we merge the graphs at different resolutions, combining any nodes that sit at the same coordinates into one node.
Note that this is possible due to how nodes align across the resolution levels.
In the \ol model only the one graph of finest resolution is used, and the 3 remaining levels disregarded.

For the \hi model the graphs of different resolution are not merged, but used as the different levels of the hierarchy.
Additional inter-level edge sets $\mupedges{l}{l+1}$ and $\mdownedges{l+1}{l}$ are then created for all adjacent levels.
Each set $\mupedges{l}{l+1}$ of upwards edges is created by connecting each node on level $l$ with the closest node on level $l+1$.
This means that each node at levels $l > 1$ will have 9 incoming edges from the level below.
The downward edges $\mdownedges{l+1}{l}$ are a copy of $\mupedges{l}{l+1}$ with the direction of each edge flipped.
The mesh graphs used in the different models are visualized in \cref{fig:3d_graphs} and the number of nodes and edges in each graph listed in \cref{tab:graph_stats}.

\begin{table}[tbp]
\centering
\caption{Number of nodes and edges in the different graphs. We count bidirectional edges twice, as these result in two messages being passed in \gls{GNN} layers (one in each direction). Grid2Mesh and Mesh2Grid describe only the edge sets, since these connect the existing grid and mesh nodes.}
\label{tab:graph_stats}
\begin{tabular}{@{}lS[table-format=4.0]S[table-format=6.0]@{}}
\toprule
 & \textbf{Nodes} & \textbf{Edges} \\ \midrule
\hi mesh graph & 7380 & 72358 \\
\ms mesh graph & 6561 & 57616 \\
\ol mesh graph & 6561 & 51520 \\ \midrule
Grid2Mesh & {--} & 100656 \\
Mesh2Grid & {--} & 255136 \\ \bottomrule
\end{tabular}
\end{table}

\subsection{Connecting the Grid and Mesh}
The grid nodes are connected to the mesh (only the bottom level in \hi) through the edge sets $\gtmedges$ and $\mtgedges$.
To form $\gtmedges$, each grid node is connected to all mesh nodes closer than $0.67 \meshdist$, where $\meshdist$ is the distance between mesh nodes at the finest resolution.
All distances are 2-dimensional euclidean, computed in the \gls{MEPS} Lambert projection coordinates.
The set $\mtgedges$ is constructed by iterating over the grid nodes, and at each creating edges from the 4 closest mesh nodes to the node in the grid.
\section{Additional Results}
\label{sec:additional_res}
In this appendix we showcase additional results from our experiments with the \gls{MEPS} data.

\subsection{Per-Variable Error}
In the subplots of \cref{fig:forecast_errors} we show the forecasting error for all 17 variables at different lead times.
Note that some variables show clear oscillations in the error, which can be explained by the day-night cycle. 
As all forecasts in the dataset were started at either 00 or 12 UTC these cycles do not average out completely in our plots.

\subsection{Example Forecasts}
\label{sec:example_predictions}
Example forecasts for all variables and models are shown in \crefrange{fig:example_pred_pres_0g}{fig:example_pred_z_1000}, where we plot predictions at lead times 15, 33 and 57 \si{\hour}.
Also the ground truth forecast from the \gls{MEPS} system is included for comparison.
The boundary area, where forcing is applied at each time step, is shown as a faded border in each plot.

\subsection{Spatial Distribution of Error}
\Cref{fig:spatial_errors} shows the spatial distribution of the error, as measured by the mean test loss computed for each grid node.
The errors in the figure come from the \hi model, but corresponding plots for the other models show similar patterns.
As expected, the error increases towards the center of the area and away from the boundary.
Information from the boundary forcing is highly useful for grid nodes close to the boundary.
To utilize this information in the center of the area, the model has to simulate physical processes over multiple time steps, which is naturally a tougher problem.
We also note that the topography has a large impact.
Errors are generally higher in areas with large gradients, such as in the Scandinavian Mountains.
As atmospheric processes are impacted by topological features this difficulty is inherent to the problem and not a unique issue of our \gls{NeurWP} method \cite{arome_metcoop, fundamentals_of_nwp}.

\newcommand{\errorsubfignamed}[3][no_legend]{%
    \begin{subfigure}[b]{0.48\textwidth}
        \centering
        \includegraphics[width=\textwidth]{graphics/error_plots/\detokenize{#1}//rmse_\detokenize{#2}.pdf}
        \caption{\wvar{\detokenize{#3}}}
        \label{fig:error_\detokenize{#2}}
    \end{subfigure}%
}
\newcommand{\errorsubfig}[2][no_legend]{\errorsubfignamed[#1]{#2}{#2}}
\newcommand{\errorplothspace}{\hspace{0.04\textwidth}}
\newcommand{\errorplotvspace}{\vspace{1em}}

\begin{figure}[b]
\vspace{-20em} %
    \errorsubfig[legend]{pres_0g}%
    \errorplothspace{}%
    \errorsubfig{pres_0s}\errorplotvspace{}
    \errorsubfignamed{nlwrs_0}{nlwrs}%
    \errorplothspace{}%
    \errorsubfignamed{nswrs_0}{nswrs}\errorplotvspace{}
    \errorsubfig{r_2}%
    \errorplothspace{}%
    \errorsubfig{r_65}%
\end{figure}
\begin{figure}[tbp]\ContinuedFloat
    \errorsubfig[legend]{t_2}%
    \errorplothspace{}%
    \errorsubfig{t_65}\errorplotvspace{}
    \errorsubfig{t_500}%
    \errorplothspace{}%
    \errorsubfig{t_850}\errorplotvspace{}
    \errorsubfig{u_65}%
    \errorplothspace{}%
    \errorsubfig{v_65}\errorplotvspace{}
    \errorsubfig{u_850}%
    \errorplothspace{}%
    \errorsubfig{v_850}%
\end{figure}
\begin{figure}[tbp]
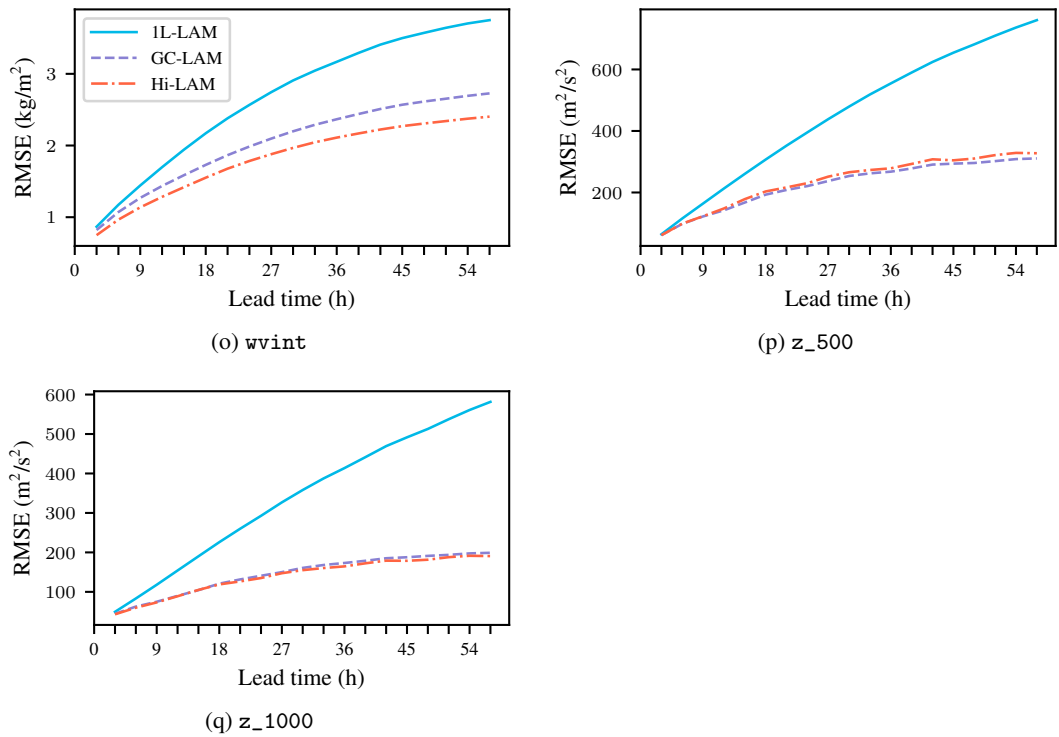
\ContinuedFloat
    \errorsubfignamed[legend]{wvint_0}{wvint}%
    \errorplothspace{}%
    \errorsubfig{z_500}\errorplotvspace{}
    \errorsubfig{z_1000}\errorplotvspace{}
    \caption{\gls{RMSE} for all models and variables, evaluated at each lead time up to 57 h.}
    \label{fig:forecast_errors}
\end{figure}

\newcommand{\expredfignamed}[2]{
    \begin{figure}[p]
        \centering
        \includegraphics[width=\linewidth]{graphics/prediction_plots/\detokenize{#1}_pred.pdf}
        \caption{Example model forecasts for \wvar{\detokenize{#2}} at lead times 15, 33 and 57 \si{\hour}.}
        \label{fig:example_pred_\detokenize{#1}}
    \end{figure}
}
\newcommand{\expredfig}[1]{\expredfignamed{#1}{#1}}
\expredfig{pres_0g}
\expredfig{pres_0s}
\expredfignamed{nlwrs_0}{nlwrs}
\expredfignamed{nswrs_0}{nswrs}
\expredfig{r_2}
\expredfig{r_65}
\expredfig{t_2}
\expredfig{t_65}
\expredfig{t_500}
\expredfig{t_850}
\expredfig{u_65}
\expredfig{v_65}
\expredfig{u_850}
\expredfig{v_850}
\expredfignamed{wvint_0}{wvint}
\expredfig{z_500}
\expredfig{z_1000}

\newcommand{\spatialsubfig}[2]{
\begin{subfigure}[b]{0.49\textwidth}
     \centering
     \includegraphics[width=\textwidth]{#1}
     \caption{\qty{#2}{\hour}}
 \end{subfigure}
}
\begin{figure}[tbp]
     \centering
     \spatialsubfig{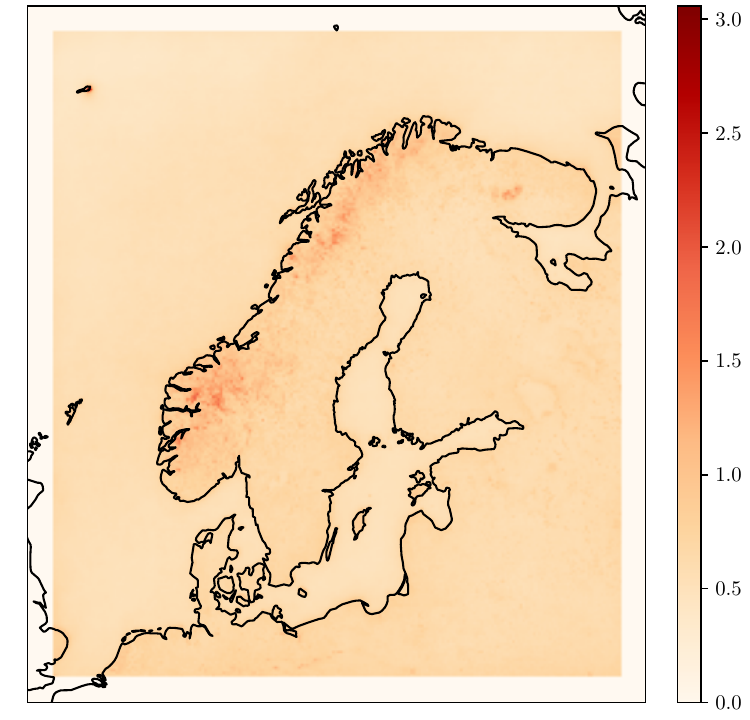}{3}
     \spatialsubfig{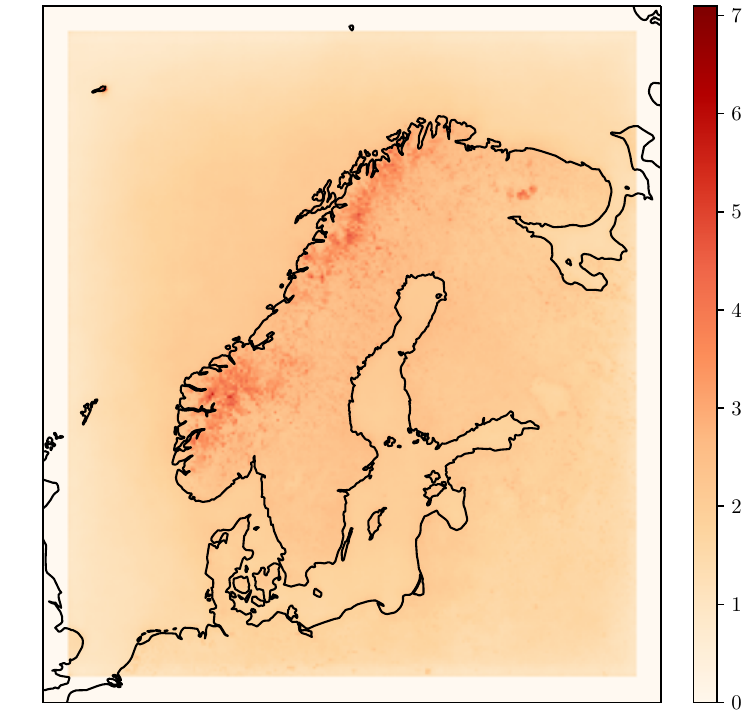}{15}
     \spatialsubfig{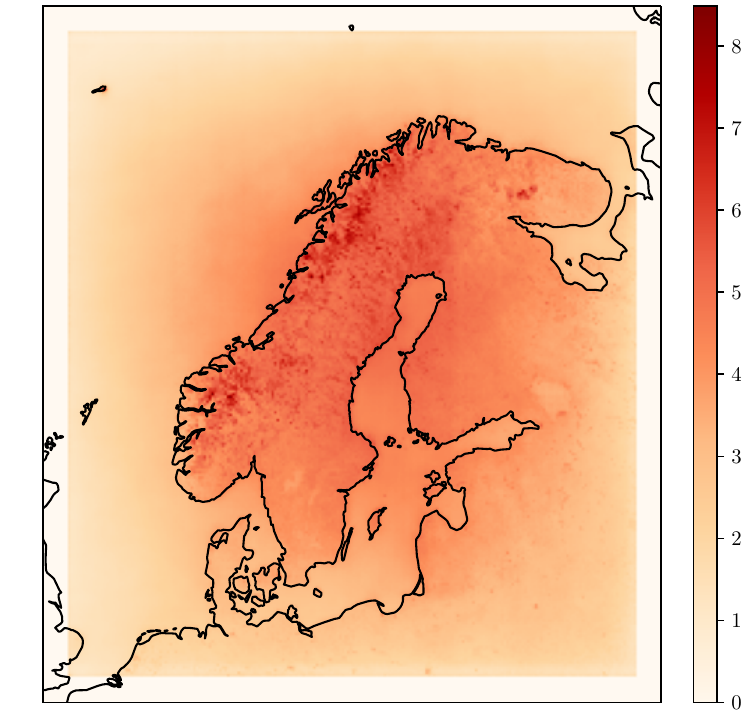}{30}
     \spatialsubfig{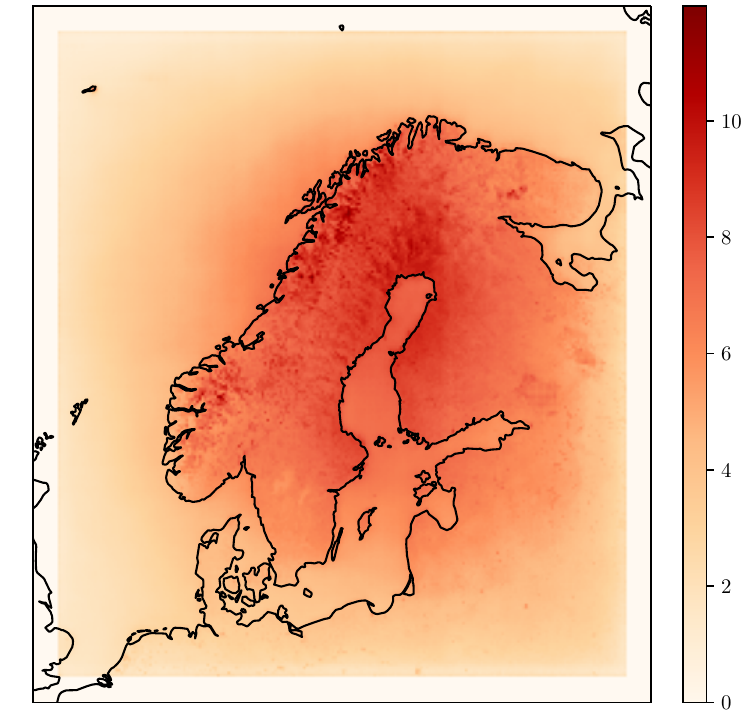}{57}
    \caption{Mean test loss of the \hi model, evaluated separately at all grid points and at different lead times. 
    Note that the color map is not shared.}
    \label{fig:spatial_errors}
\end{figure}

\bibliographyapp{references}
\bibliographystyleapp{unsrtnat}

\end{document}